\documentclass{article}

\usepackage[authoryear]{natbib}
\setcitestyle{authoryear,round}

\usepackage[utf8]{inputenc} %
\usepackage[T1]{fontenc}    %
\usepackage{lmodern}
\usepackage{hyperref}
\usepackage{url}            %
\usepackage{amsfonts}       %
\usepackage{nicefrac}       %
\usepackage{microtype}      %
\usepackage{xcolor}         %

\usepackage{wrapfig}
\usepackage{graphicx}
\usepackage{subcaption}

\usepackage{latexsym}
\usepackage{enumitem}

\usepackage{fullpage}

\usepackage[normalem]{ulem}
\usepackage[skins]{tcolorbox}

\usepackage{booktabs}
\usepackage{multirow}  
\usepackage{array}      
\usepackage{subcaption}
\usepackage{makecell}
\usepackage{tabularx}

\usepackage{longtable}
\usepackage{dblfloatfix}

\usepackage{titletoc}

\usepackage{footmisc}

\usepackage{amsmath}
\usepackage{amssymb}
\usepackage{mathtools}
\usepackage{amsthm}
\usepackage{bm}

\usepackage{xspace}
\usepackage{etoc}

\title{Policies Permitting LLM Use for Polishing Peer Reviews \\ Are Currently Not Enforceable}

\newcommand{\huggingface}{\raisebox{-1.5pt}{\includegraphics[height=1.05em]{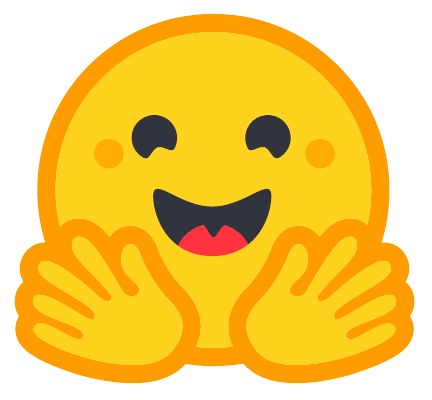}}\xspace}
\newcommand{\github}{\raisebox{-1.5pt}{\includegraphics[height=1.05em]{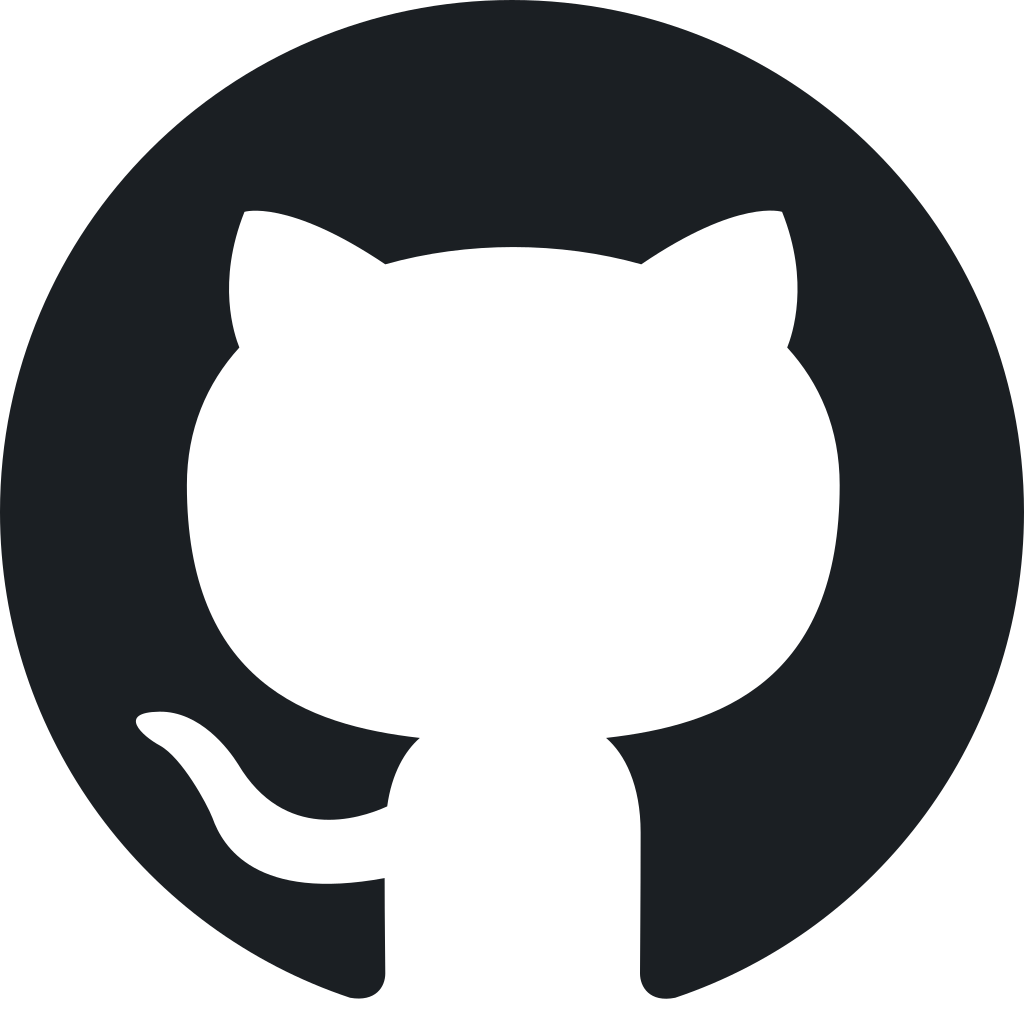}}\xspace}

\date{} 

\author{
    \vspace{0.5em} 
    \textbf{Rounak Saha$^{\spadesuit}$} \quad
    \textbf{Gurusha Juneja$^{\diamondsuit *}$} \quad
    \textbf{Dayita Chaudhuri$^{\spadesuit *}$} \quad
    \textbf{Naveeja Sajeevan$^{\spadesuit *}$} \quad \\
    \textbf{Nihar B Shah$^{\heartsuit}$} \quad
    \textbf{Danish Pruthi$^{\spadesuit}$} \quad 
    \vspace{0.4em} \\
    \textnormal{$^{\spadesuit}$Indian Institute of Science \quad $^{\diamondsuit}$}University of California, Santa Barbara  \\ \textnormal{$^{\heartsuit}$Carnegie Mellon University}
}

\usepackage{soul,xcolor}

\definecolor{vocabcolor}{RGB}{255, 245, 200}   
\definecolor{structcolor}{RGB}{220, 240, 255} 
\definecolor{contentcolor}{RGB}{230, 255, 230}   

\newcommand{\vocab}[1]{\sethlcolor{vocabcolor}\hl{#1}}
\newcommand{\struct}[1]{\sethlcolor{structcolor}\hl{#1}}
\newcommand{\content}[1]{\sethlcolor{contentcolor}\hl{#1}}

\definecolor{goodgreen}{RGB}{34,139,34}
\definecolor{badred}{RGB}{255,0,0}
\definecolor{darkred}{RGB}{200,0,0}
\definecolor{lightgray}{gray}{0.9}
\definecolor{lightgreen}{HTML}{EDF8FB}
\definecolor{mediumgreen}{HTML}{66C2A4}
\definecolor{darkgreen}{HTML}{006D2C}

\newcommand{\quotecode}[1]{\texttt{#1}}
\newcommand{\hltred}[1]{\textcolor{darkred}{\bm{#1}}}

\newcommand{\base}{easy}
\newcommand{\hard}{hard}
\newcommand{\Base}{Easy}
\newcommand{\Hard}{Hard}
\newcommand{\splitsuffix}{subset}

\newcommand{\basesplit}{\base \space \splitsuffix}
\newcommand{\hardsplit}{\hard \space \splitsuffix}
\newcommand{\Basesplit}{\Base \space \splitsuffix}
\newcommand{\Hardsplit}{\Hard \space \splitsuffix}

\newcommand{\lone}{AI-BP}
\newcommand{\ltwo}{AI-EP}
\newcommand{\lthree}{AI-HI}
\newcommand{\lfour}{H-AI}
\newcommand{\lfive}{H}
\newcommand{\levelone}{AI-generated with Basic Prompts}
\newcommand{\leveltwo}{AI-generated with Elaborate Prompts}
\newcommand{\levelthree}{AI-generated with Human Input}
\newcommand{\levelfour}{Human-written AI-polished}
\newcommand{\levelfive}{Completely Human}
\newcommand{\fullyAI}{AI$^*$}
\newcommand{\fullyHuman}{H$^*$}

\begin{document}

\maketitle
{\renewcommand{\thefootnote}{}%
\footnotetext{$^{*}$Equal contribution. Correspondence to: Rounak Saha $\langle$\texttt{rounaksaha@iisc.ac.in}$\rangle$, Danish Pruthi $\langle$\texttt{danishp@iisc.ac.in}$\rangle$.\par\smallskip
\textit{Published in the 43\textsuperscript{rd} International Conference on Machine Learning, Seoul, South Korea, 2026.}}}

\begin{abstract}
    A number of scientific conferences and journals 
    have recently enacted policies that 
    prohibit LLM usage by peer reviewers, except for polishing, paraphrasing, and grammar correction of otherwise human-written reviews. But, are these policies enforceable? 
    To answer this question, 
    we assemble a dataset of peer reviews simulating multiple levels of human-AI collaboration, and evaluate five state-of-the-art detectors, including two commercial systems. Our analysis shows that all detectors misclassify a non-trivial fraction of LLM-polished reviews as AI-generated, thereby risking false accusations of academic misconduct. We further investigate whether peer-review-specific signals, including access to the paper manuscript and the constrained domain of scientific writing, can be leveraged to improve detection. While incorporating such signals yields measurable gains in some settings, we identify limitations in each approach 
    and find that none meets the accuracy standards required for identifying AI use in peer reviews. 
    Importantly, our results suggest that 
    recent public estimates of 
    AI use in peer reviews through the use of AI-text detectors 
    should be interpreted with caution, 
    as current detectors misclassify 
    mixed reviews (collaborative human-AI outputs)
    as fully AI generated, 
    potentially overstating the extent of policy violations.
\end{abstract}

\begin{center}
\github \textbf{\small Code}: \texttt{\href{https://github.com/FLAIR-IISc/ai-in-peer-review}{\small https://github.com/FLAIR-IISc/ai-in-peer-review}}

\vspace{0.5em}

\huggingface \textbf{\small Data}: 
\texttt{\href{https://huggingface.co/datasets/rounaksaha12/ai-in-peer-review}{\small https://huggingface.co/datasets/rounaksaha12/ai-in-peer-review}}
\end{center}

\section{Introduction}
\label{sec:intro}
Scientific journal editors and conference program chairs routinely rely on experts to evaluate submitted manuscripts through peer review. However, there have been growing concerns lately about reviewers delegating this task to AI systems, prompting journals and conferences to adopt a range of policies governing AI use in the review process~\citep{icml2025reviewerinstructions,Wiley2025AI}. These concerns have also been a part of public discourse. A notable recent example involves the ICLR 2026 conference, whose reviews are publicly released. An AI-text detection company claimed that $21$\% of submitted reviews were AI-generated~\citep{pangram21percentage}. This claim attracted widespread attention, but was discussed and reported~\citep{naddafnature} with little scrutiny.

Two types of policies are prominently adopted by conferences and journals. One policy is to completely prohibit use of LLMs by reviewers (e.g., by~\citealp{icml2025reviewerinstructions}), and we will refer to this as \textbf{No-LLM-use policy}. A second common policy is to allow reviewers to use LLMs only for paraphrasing and grammar correction of their reviews, and where the core arguments and content of the review must be human-written (e.g., by ~\citealp{iclr2026llmpolicy}, ~\citealp{arrreviewerguidelinesgenai},
~\citealp{TaylorFrancis2026}). We collectively refer to these policies as \textbf{Polishing-only policies}. 
For instance,~\citealp{EMNLP2025}, 
citing the ACL Policy on Publication Ethics, states:

\begin{quote}
\emph{``it is acceptable to use LLMs for paraphrasing, grammatical checks and proof-reading, but not for the
content of the (meta-)reviews.''}

\hfill --- Criteria for Determining Irresponsible Reviewers, \citealp{EMNLP2025}
\end{quote}

With such policies precisely delineating 
acceptable LLM use, 
and detectors claiming to be capable of screening tens of thousands of reviews overnight, 
it might be tempting to conclude that the machinery for enforcing these policies is already in place. 
Enforcement of policies, however, 
hinges critically on the key question: 
\textit{To what extent 
can  
the extent of AI use 
in peer reviews be detected?} 
\hypertarget{polishing-only-stakes}{To put things into perspective,
at the scale of NeurIPS 2025 
(over 75,000 reviews), 
even a $0.1\%$ false positive rate  can 
translate to 75 wrongful accusations
of academic misconduct. A $3\%$ false positive rate can result in an alarmingly high 2,250 wrongful accusations. 
Therefore, any credible enforcement regime would 
demand a far lower false positive rate still.}
Yet, it remains largely unexamined
how reliable are these detectors 
in detecting AI use in peer reviews,
and how much confidence 
we should place
on their outputs.
In this work, we directly investigate these questions 
by evaluating state-of-the-art AI text detectors 
on a curated dataset of reviews 
spanning multiple levels of human-AI collaboration.

\paragraph{Data.}
We curate a dataset of $50,156$ peer reviews spanning multiple levels of human-AI collaboration. We start with {\bf human}-written reviews from the pre-ChatGPT era to ensure reviews are free of AI assistance, specifically from CoNLL 2016, ACL 2017, ICLR 2017, and NeurIPS 2013–2017, sourced from the PeerRead dataset~\citep{kang-etal-2018-dataset}. We then use popular LLMs to generate synthetic reviews for the same papers to simulate the following levels of AI-assistance:
\begin{enumerate}[label=(\arabic*),nosep]
    \item \levelone~(\textbf{\lone{}}): Reviews generated by prompting an LLM with the paper and conference reviewing guidelines;
    \item \leveltwo~(\textbf{\ltwo{}}): Reviews generated by prompting with additional tips apart from paper and official reviewing guidelines;
    \item \levelthree~(\textbf{\lthree{}}): Reviews generated by an LLM by expanding on key assessment points extracted from a human-written review, simulating minimal human input;
     \item \levelfour{}~(\textbf{\lfour{}}): Human-written reviews polished by an LLM for grammar, flow and clarity, the only form of LLM use permitted under Polishing-only policies; and
    \item \textbf{Humanized} \lone{} and \lfour{} reviews, adversarially paraphrased to evade AI detection. 
\end{enumerate}

\paragraph{Detectors evaluated.}
We evaluate five AI text detectors, comprising three open-source and zero-shot AI detectors, namely LogLikelihood~\citep{loglikelihood}, Fast-DetectGPT~\citep{fastdetectgpt}, and Binoculars~\citep{binoculars}, 
which output scalar detection scores,
as well as two commercial classifiers, 
Pangram~\citep{pangram, thai2025editlens} and GPTZero~\citep{tian2026gptzero}, which produce multi-class categorical predictions 
(``AI'', ``Mixed'', ``Human'').
For the zero-shot detectors,
we generate binary predictions 
by thresholding their output scores.
The thresholds are chosen 
to achieve a hard $0\%$ false positive rate on human-written reviews in a calibration set.

\paragraph{Potential peer-review-specific advantages.}
As opposed to general AI-text detection, the application of detecting AI use in peer reviews 
presents unique opportunities. 
For instance, 
the paper manuscript (under review) 
is likely a part of the prompt used to generate the review.
Most open-source detectors like LogLikelihood, DetectGPT, Fast-DetectGPT and Binoculars rely on perplexity or 
surprisal-based metrics, which can be naturally extended to incorporate paper context by computing the likelihood of a review conditioned on the manuscript. 
Another way to incorporate paper context is to first generate a set of LLM-written reference reviews for the manuscript, and then use similarity between a candidate review and these references as a detection signal~\citep{yu2025your}.
Secondly, 
peer reviews follow certain 
writing norms and styles, 
and the 
content is scientific in nature, 
which makes supervised training 
on domain-specific examples
a promising candidate.
This raises the question:  
\textit{Does the presence of peer-review-specific signals 
help overcome 
the limitations of fine-grained AI detection in peer reviews?}

\paragraph{Main findings.} 
We summarize our {key findings} below:
\begin{itemize}
    \item Pangram and GPTZero successfully identify $98.3\%$ and $95.8\%$ of fully AI-generated reviews (\lone{} and \ltwo{}), respectively. 
    For reviews that are fully human written, Pangram flags none of them as ``AI'' or ``Mixed'', while GPTZero flags $1\%$ of such reviews as ``AI.''\footnote{The human reviews are drawn from the PeerRead dataset, which might have appeared in the training data of Pangram and GPTZero. Therefore, these results are potentially optimistic estimates. More on this in \hyperlink{contamination-note}{the note on contamination} (Sec.~\ref{subsec:off-the-shelf-det-perf})}
    \item All detectors struggle with the \levelfour{} (\lfour{}) reviews. These reviews are compliant under a Polishing-only policy. 
    However, Pangram and GPTZero have a non-trivially high false positive rate, classifying     
    $3.1\%$ and $3.4\%$ of such reviews as ``AI'' (not even ``Mixed''), respectively. 
    Further experiments on other forms of 
    real-world AI-assisted polishing behaviour 
    such as Grammarly-style edits, 
    LLM-based translation and personalized prompts 
    reveal similar non-trivial false positive rates across all but Grammarly-style edits (Appendix~\ref{appdx:addtnl-polishing-variants}).

    \item Zero-shot open-source detectors perform notably worse compared to the proprietary ones. The best-performing of them in our experiments, Fast-DetectGPT, can identify only $70\%$ of fully AI-generated (\lone{}, \ltwo{}) reviews with false positive rates of $4.6\%$ and $0.2\%$ on AI-polished and human written reviews respectively. Taking advantage of the availability of the paper manuscript under review, when the perplexities are computed conditioned on the paper manuscript, their ability to detect \lone{}, \ltwo{}, \lthree{} reviews improves, but at the cost of even higher false positive rates.

    \item Under humanization, Pangram and GPTZero 
    continue to flag the vast majority 
    of humanized \lone{} reviews as either ``AI'' or ``Mixed''
    rather than ``Human'' ($92.8\%$ and $86.7\%$, respectively),
    and so remain useful in determining
    whether some form of AI-assistance was involved.
    However, surprisingly, \lfour{} reviews tend to be flagged as
    ``AI'' more frequently if they are humanized.

    \item To leverage peer-review specific signals, we measure similarity of a candidate review with several AI-generated reference reviews (given we have access to the paper manuscript). We find that these similarity scores yield different distributions across varying levels of AI assistance, at an aggregate level.  
    However, these distributions overlap substantially and many individual AI-polished reviews receive similarity scores indistinguishable from those of fully AI-generated reviews.
    Therefore, this approach can not be used for reliable review-level detection.

    \item The narrow writing style and scientific content of peer reviews, while a natural motivation for supervised training, does not lead to much added reliability in detection. Supervised classifiers based on stylometric features and dense embeddings fail to generalize beyond the LLMs whose reviews they were trained on.

    \item 
    Finally, we offer a practical note for reviewers using LLMs to polish their drafts. Uploading the submitted paper, or omitting explicit content-preservation instructions in the polishing prompt, can lead LLMs to introduce new content in the resulting review rather than just polishing it. Reviews polished using these prompts are classified as ``AI'' significantly more often. Reviewers should therefore phrase their prompts carefully and verify the `polished' output before submission. 
\end{itemize}

Overall, we find 
that 
neither 
off-the-shelf detectors 
nor strategies that leverage peer-review-specific advantages
lead to systems that can 
consistently classify LLM-polished reviews
into the same category as human-written or mixed ones.
This corroborates
recent findings on the challenges of 
detecting mixed-authorship text in non peer-review data~\citep{almostaialmosthuman,llmascoauthor}.
This finding has direct policy implications: it 
is infeasible to enforce
any policy 
that permits LLM usage selectively for improving flow, clarity or grammar correction. 
In contrast, 
under a blanket No-LLM-use policy, 
AI-polished reviews could be misclassified as human-written, 
in which case these (supposedly minor) violations would go undetected.  
Beyond enforcement of policies, our findings
carry implications on how 
detection performance should be
interpreted and communicated. 
For instance, the recent estimates 
about $21\%$ of peer reviews in ICLR 2026 being fully AI-generated by \cite{pangram21percentage} 
may not be accurate, as these estimates likely include 
reviews written with only partial AI involvement.

\paragraph{AI detection on full papers and a note on NeurIPS PPT desk rejections.} Recently, NeurIPS 2026 Position Paper Track desk-rejected 178 submissions ($18.4\%$) 
for substantial AI use measured in terms of 
the fraction of a paper's text windows flagged as AI-generated by Pangram~\citep{neurips2026aigenerated}.
Our results, though relevant to this discussion, must be read with care. 
The reviews in our dataset are short ($\sim350$ words) and mostly fit within a single Pangram window.
Therefore, our $\sim3\%$ false positive rate on AI-polished (\lfour{}) reviews is best understood as a per-window rate, not a per-paper one. 
Full papers span many windows, and the rejections targeted papers where a large majority of windows ($50\%$ in the most strict case) were flagged.
If window-level false positive rates were independent, 
then it is improbable that a paper which is only polished using LLMs would have more than $50\%$ of their windows flagged as fully AI-generated by Pangram.
So our findings need not contradict these decisions, especially given the additional safeguards adopted (smaller 100-word windows, author AI-use declarations, and submission-pattern checks). 
But one key caveat is that this reasoning assumes independence across windows from the same paper. 
An author's individual writing style may induce correlated errors across windows, and our single-window reviews provide no evidence for or against such correlation. Per-window false positive rates should therefore not be extrapolated to whole papers in either direction.

\section{Related Work}
\label{sec:related}

Recent OpenAI and Anthropic reports studying ChatGPT \citep{howpeopleusechatgpt} and Claude \citep{howpeopleuseclaude} usage patterns indicate that people frequently delegate 
tasks that require analyzing, summarizing and reviewing information, in addition to creative writing, to AI. 
Writing is a dominant use case across both platforms.
Several recent studies attempt to estimate the amount of AI participation in peer review writing for scientific conferences.  
\cite{corpuslevelmonitoring} 
estimate that
an approximate $6.5$-$16.9\%$ 
of all the peer review text from 
ICLR 2024, NeurIPS 2023, CoRL 2023 and EMNLP 2023 
are AI-modified beyond simple writing updates.  
Detecting individual AI-generated reviews remains challenging. 
Recently, ICML 2026~\citep{ICMLblogLLMreview2026} adopted the strategy proposed by~\citet{rao2025detecting} whereby it
injected prompts in white text in
manuscript PDFs
instructing the AI models to embed a covert watermark in the generated review which can later be detected with strong statistical guarantees on family wise error rates.
A broader discussion on the different enforcement mechanisms popularly adopted by conferences is provided in Appendix~\ref{appdx:enforcement-landscape}.

Current post-hoc AI text detectors broadly fall into two categories: (a) supervised classifiers such as OpenAI's AI Text Classifier~\citep{openai2023classifier} and RADAR~\citep{radar}, that train neural networks on human-AI parallel datasets, and (b) zero-shot methods that leverage statistical signatures of AI-generated text without training such as perplexity~\citep{loglikelihood}, perplexity curvature 
(DetectGPT,~\citealp{detectgpt} 
and 
Fast-DetectGPT,~\citealp{fastdetectgpt}), 
and cross-perplexity (Binoculars,~\citealp{binoculars}).
However, these detectors often fail to identify AI-generated reviews, with evaluations showing that up to $40\%$ of purely AI-generated reviews are misclassified as human-written~\citep{yu2025your}.
Our study finds that more recent proprietary AI detectors like Pangram~\citep{pangram, thai2025editlens} and GPTZero~\citep{tian2026gptzero}, 
trained on 
very large scale datasets
with human content and their AI-generated mirrors,
have substantially improved detection of fully AI-generated content, 
a finding corroborated by independent benchmarks \citep{jabarian2025artificial}.
Several studies estimating 
the prevalence of AI involvement in peer reviews and academic writing rely on the outputs of these detectors~\citep{aireviewlottery,elazar2026llmgeneratedhumanwrittencomparingreview}.
Meanwhile, ChatGPT usage data \citep{howpeopleusechatgpt}
indicates that nearly two-thirds of writing-related interactions in the platform involve modifying user-provided texts rather than generating new text from scratch. 
This suggests that AI involvement in peer reviews is better characterized as a spectrum, ranging from light stylistic polishing to 
completely generating reviews, 
which we model in our 
evaluation. 

Prior work has documented 
misclassification of AI-polished text~\citep{almostaialmosthuman, llmascoauthor}
as entirely AI-generated, 
but peer review presents 
high stakes policy implications 
as well as structural advantages 
to the problem of AI text detection.
Our work evaluates promising directions for fine-grained AI detection that have emerged recently: for instance, \citet{thai2025editlens} find that the distance from parallel human-author content can serve as an effective supervision signal for training 
a model that can quantify the extent of AI editing involved. This model, called EditLens, is commercially available as Pangram 3.0. 
Additionally, our work also explores whether unique opportunities in the peer review setup (e.g. availability of the paper under review and domain specificity) 
can help
circumvent concerns of false positives on AI-polished human content and discusses their implications on the enforceability of Polishing-only policies in peer reviews.

Finally, we also clarify that our position is not against the use of LLMs in reviewing. In fact, LLMs can be particularly useful in ensuring scientific rigor (\citealp{liu2023reviewergpt, xi2025flaws};~\citealp[Sections 10.1.2 and \S11.3.2]{shah2025survey}) although LLM reviewers need to be evaluated appropriately (\citealp{baumann2026stopautomatingpeerreview};~\citealp[Part 1]{shah2025aipeerreview}).
Our work simply aims to evaluate current policies adopted by journals and conferences, and alongside, provide more evidence to supplement broader discourse on this topic.

\section{Evaluation Framework}
\label{sec:experimental_setup}

In this section, we outline the creation of the peer-review dataset and the metrics used to evaluate various detectors.

\textbf{Dataset of human-written reviews.} 
Our study requires peer reviews authored without AI assistance.
To be certain that reviews are human written, 
we use 
pre-$2020$ review data (before GPT-3's release).
The PeerRead dataset~\citep{kang-etal-2018-dataset} comprises $14,000$ paper drafts and $10,000+$ expert reviews from conferences including NeurIPS, ACL, ICLR, and CoNLL, all from 2017 or earlier. We work with a subset of $1086$ papers ($500$ papers from NeurIPS and full set from ACL, ICLR, CoNLL) and $3499$ corresponding human-written reviews for papers from these venues 
(Table~\ref{tab:dataset_distribution}).

\begin{wraptable}[10]{r}{0.42\textwidth}
\vspace{-1em}
\caption{\centering Distribution of papers and human-written reviews.
}
\label{tab:dataset_distribution}
\centering
\footnotesize
\setlength{\tabcolsep}{2.5pt}
\begin{tabular}{lcc}
\toprule
\textbf{Conference} & \makecell{\textbf{Papers}} & \makecell{\textbf{Human Reviews}}\\
\midrule
CONLL 2016 & $22$ & $39$ \\
ACL 2017 & $137$ & $275$ \\
ICLR 2017 & $427$ & $1303$ \\
NeurIPS 2013-17 & $500$ & $1882$ \\
\midrule
\textbf{Total} & $1086$ & $3499$ \\
\bottomrule
\end{tabular}
\end{wraptable}

\textbf{Reviews generated with different levels of Human involvement.} 
We construct a dataset of reviews that systematically vary in their degree of human involvement. Specifically, we generate reviews at five levels of human involvement, ranging from completely AI-generated reviews to fully human-authored reviews. While the exact forms of AI assistance adopted by reviewers in practice remain undetermined, these levels are constructed to simulate plausible scenarios. 
In the descriptions that follow, we motivate each level and specify the inputs provided to the LLM.

\begin{itemize}[leftmargin=2em,itemsep=1pt]

    \item \textbf{\levelone{} (\lone{}):} 
    This level captures the most disengaged reviewer.
    To generate \lone{} reviews, we prompt an LLM with just the paper and the official reviewing guidelines of the respective conference.
    
    \item \textbf{\leveltwo{} (\ltwo{}):} 
    This level simulates a lazy reviewer who is just resourceful enough to grab general reviewing advice and best practices 
    from the internet and incorporate them into the prompt,
    but does not contribute any original opinion or content of their own. 
    We aggregate best practices from
    five conference-issued reviewer best-practice documents
    (e.g., ACL 2017 Last Minute Reviewing Advice;\footnote{\href{https://acl2017.wordpress.com/2017/02/23/last-minute-reviewing-advice/}{ACL 2017 last minute reviewing advice}} full list in Appendix~\ref{appdx:prompts}). Alongside the paper and reviewing guidelines, these 
    conference-issued reviewer best practices
    are included in the prompt.
    
    \item \textbf{\levelthree{} (\lthree{}):} 
    This level represents the case where reviewers identify key assessment points after skimming the paper and instruct an LLM to elaborate on them. We simulate this via a two-step process: first, an LLM extracts key assessment points from a human-written review; these key points are then included 
    in the prompt for an LLM to elaborate upon. We also include the paper text and reviewer guidelines in the prompt.
    
    \item \textbf{\levelfour{} (\lfour{}):} 
    This is the only acceptable form of LLM use under ``Polishing-only'' policies. 
    Here, 
    the LLM is given just a human-written review, without the paper manuscript or reviewing guidelines, and is instructed to improve grammar, flow and clarity while preserving structure, meaning and technical content. 
    In addition to explicitly including content-preservation instructions in the prompt, we filter 
    out reviews whose length exceeds $1.25$ times that of the original human review to limit the LLM's involvement to polishing only (more on this in Section~\ref{subsec:off-the-shelf-det-perf}).

    \item \textbf{\levelfive{} (\lfive{}):} 
    This level contains the original human-authored peer reviews from the PeerRead dataset.

\end{itemize}

\begin{figure}[t]
    \centering
    \includegraphics[width=0.7\textwidth]{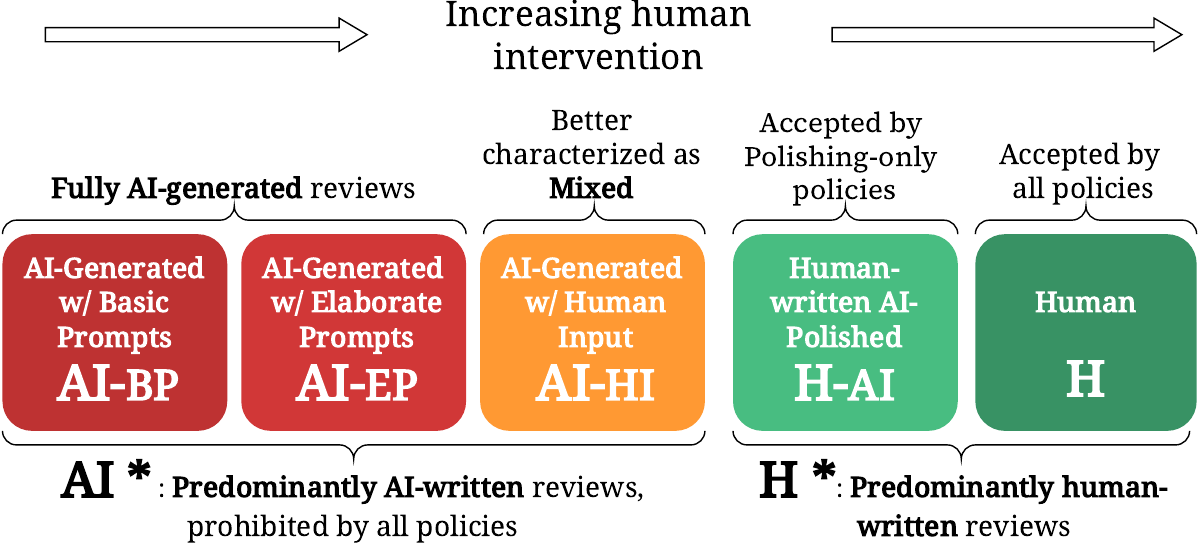}
    \caption{
    Levels of AI-assistance.
    }
    \label{fig:levels_explained}
\end{figure}

An overview of the above levels is available in Figure~\ref{fig:levels_explained}.
In the rest of the paper, we use the term \textbf{\fullyAI{}} to refer to the first three levels combined (\lone{}, \ltwo{} and \lthree{}).
\lone{} and \ltwo{} are clear violations of Polishing-only policy, while \lfour{} is permitted.
\hypertarget{ai-hi-discussion}{One might argue that for \lthree{} reviews, the LLM's involvement is limited to stitching together the human-written key assessment points into prose and hence can be considered as AI-polished.} However, unlike \lfour{} reviews, the \lthree{} prompt explicitly instructs the LLM to elaborate on these key points, going beyond merely polishing the reviews. The LLM also has access to the paper manuscript to expand the key points. On average, the length of the \lthree{} reviews (679 words) is roughly 4 times the length of keypoints (163 words) used to generate them. Though these reviews are best classified as ``mixed’’, while reporting performance of detectors, we consider them under the ``positive’’ class (violation under Polishing-only policy). For detectors whose predictions include a ``mixed’’ label, we separately report the full distribution of predicted labels.
To summarize, while reporting performance of detectors concisely in terms of true and false positive rates, \fullyAI{} reviews are considered ``positives’’ and the rest, together denoted as \textbf{\fullyHuman{}}, are considered ``negatives.’’

For each paper, we generate reviews using GPT-4o and Llama-3.3-70B-Instruct models, with one prompt per level. This results in $18,122$ AI-generated reviews, which we refer to as the \textbf{\basesplit{}}.
The \basesplit{} provides higher coverage in terms of number of papers, but limited diversity in the choice of models generating the reviews. Therefore, additionally, for a subset of $158$ papers ($18$ from CoNLL 2016 and $20$ from each of the remaining conferences), we generate reviews using GPT-5, Gemini-2.5-pro, Gemma-3-27b-it, Qwen-3-30B-thinking and Llama-3.1-70B-instruct. For this subset, we employ at least four distinct prompts per level, resulting in $26,535$
AI-generated reviews, which we refer to as the \textbf{\hardsplit{}} (see Appendix~\ref{appdx:prompts} for full set of prompts).
With newer models and diverse prompts, the \hardsplit{} more closely captures AI assistance in the wild and yields a more challenging evaluation. Combined with $3499$ human-written reviews, our dataset contains $48,156$ 
including both the \base{} and \hard{} subsets.

\textbf{Humanized Reviews.} 
``Humanization''  refers to the paraphrasing of AI-generated text, with the deliberate aim of mimicking human writing 
to evade detection~\citep{Shi2023RedTL,Wang2024HumanizingTM,masrour-etal-2025-damage}. Numerous automated tools are publicly available and commonly used for this purpose. 
Note that this adversarial paraphrasing is different from 
the 
collaborative
paraphrasing scenario that \lfour{} reviews represent, where a reviewer uses AI to refine their draft. 
We employ Undetectable AI,\footnote{\url{https://undetectable.ai/}} 
one of the most widely-used 
commercial AI humanizers
with an API, 
to humanize reviews from two key levels: 
\lone{}
and 
\lfour{}. 
Humanizing AI reviews 
simulates the scenario of a lazy reviewer trying to present a fully AI-generated review as their own. We also consider a second plausible scenario, where an overly cautious reviewer, who has polished their draft with AI (\lfour{} review), chooses to pass it through a humanization tool to minimize the risk 
of detection. 
Including $2000$ humanized reviews, the size of our dataset becomes $50,156$.

\textbf{Metrics and Calibration.} To evaluate a detector, we report the fraction of reviews 
predicted
as AI-generated (``positive'' class) at each level. 
This corresponds to true positive rate (TPR) for \fullyAI{} reviews and
false positive rate (FPR) for \fullyHuman{} reviews
following our earlier discussions
outlining the rationale of deciding
which levels constitute violation of
Polishing-only policy.
For open-source zero-shot detectors that output scalar detection scores, we use a calibration set of 
human-written reviews from NeurIPS 2013-2015 to adjust the thresholds for the ``positive'' class. The thresholds are chosen to ensure $0\%$ FPR on the calibration set. We evaluate these detectors on reviews of conferences after 2016. The rationale is to simulate a situation where ground truth from past conferences is used to calibrate the detectors. For fair comparison with other detectors, we also include examples from NeurIPS 2013-2015 in the evaluation set, ensuring none of them belong to the same paper as any review in the calibration set (Table~\ref{tab:detector_scores_on_test_data}).
For Pangram and GPTZero,
which produce three-class predictions 
(``AI'', ``Mixed'', ``Human''), 
we report TPR and FPR in Table~\ref{tab:detector_scores_on_test_data} 
by treating only reviews labeled ``AI'' as the positive class, 
collapsing ``Mixed'' with ``Human''. 
This conservative choice minimizes false positives on policy-compliant reviews. 
Additionally, we separately examine 
the full distribution of predicted labels across all three classes, 
reported as confusion matrices in Figure~\ref{fig:conf-mat}, 
to provide a more complete picture of detector performance.

\section{Experiments \& Results}
\label{sec:experimental_results}

In this section we discuss the experiments we conducted and enumerate our findings.

\subsection{Can off-the-shelf AI detectors distinguish between different levels of AI assistance in peer reviews?}
\label{subsec:off-the-shelf-det-perf}

We evaluate
a set of $5$ detectors
on our review dataset,
selected to represent the landscape of currently available AI detection tools. This includes $3$ open-source and $2$ commercial detectors. Detailed descriptions and hyperparameter settings for these detectors are provided in Appendix~\ref{appdx:auto-detect-descp}.

\begin{figure}[t]
    \centering
    \includegraphics[width=0.8\textwidth]{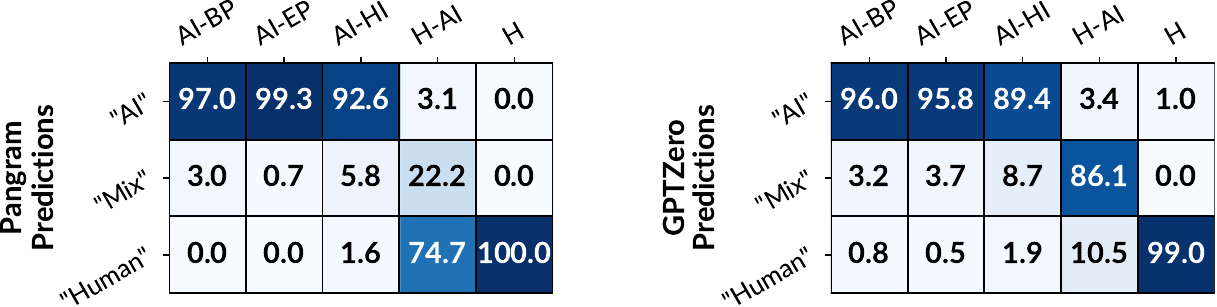}
    \caption{
    Confusion matrices denoting \% of reviews classified as ``AI'', ``Mixed'', ``Human'' by Pangram $\&$ GPTZero on \hardsplit{}.
    }
    \label{fig:conf-mat}
\end{figure}

We find that \textbf{Pangram correctly flags more than $90\%$ of AI-generated reviews with minimal to no human input (\fullyAI{} reviews), while maintaining $0\%$ FPR on human reviews.}  
Pangram and GPTZero both return fine-grained labels of AI involvement, ``AI'', ``Mixed'' or ``Human''. 
Figure~\ref{fig:conf-mat} presents the distribution of reviews in each level across these labels on the \hardsplit{} (similar plots for the \basesplit{} are provided in Appendix~\ref{appdx:easy-split-conf-mat}).
In the \hardsplit{}, Pangram consistently predicts more than $97\%$ of \fullyAI{} reviews as ``AI'', except for \lthree{} reviews for which it classifies $92.6\%$ as ``AI'' and $5.8\%$ as ``Mixed''. GPTZero also classifies high fraction (greater than $97\%$) of \fullyAI{} reviews as either ``AI'' or ``Mixed'', but unlike Pangram, has non-zero FPR on human-written reviews. 
In contrast, \textbf{open-source zero-shot AI detectors cannot detect \fullyAI{} reviews generated with newer models and varied prompts.} 
Fast-DetectGPT, the best performing open-source detector in our experiments, has near perfect detection rates, on par with commercial detectors on the \basesplit{}. However, in the \hardsplit{}, which is characterized by more modern LLMs and varied prompts, its ability to detect \fullyAI{} reviews drops sharply into the low-$70\%$ range, 
limiting their practical utility for policy enforcement.

\begin{table*}[t]
\captionof{table}{
Performance of detectors with and without paper context. 
Pangram and GPTZero are context-agnostic by design.  
We note that in many cases, zero-shot detectors {\textcolor{blue}{with paper context}} yield marginal improvements over the no-context case
($^*$ indicates statistically significant difference, more details in  Appendix~\ref{appdx:statistical_significance}).
Detection performance of commercial systems, Pangram and GPTZero, is significantly better.
However, both of them classify a \textbf{\textcolor{darkred}{non-trivially high percentage}} of human-written but AI-polished (\lfour{}) reviews as AI-generated, even when ``Mixed'' labels are regarded as human-written (more on this in Section~\ref{subsec:off-the-shelf-det-perf} and Figure~\ref{fig:conf-mat}). {Hence, Polishing-only policies are currently not enforceable.}
}
\label{tab:detector_scores_on_test_data}
\centering
\vspace{0pt}
\small
\setlength{\tabcolsep}{3.5pt} %

\begin{tabular}{lccccccccc}
\toprule
& \multicolumn{4}{c}{\textbf{\Basesplit{}}}
& \multicolumn{4}{c}{\textbf{\Hardsplit{}}}
& \textbf{Human}\\
\cmidrule(lr){2-5} \cmidrule(lr){6-9} \cmidrule(lr){10-10}
\textbf{Method}
& \multicolumn{3}{c}{\textbf{TPR (\%)} $\uparrow$}
& \multicolumn{1}{c}{\textbf{FPR (\%)} $\downarrow$}
& \multicolumn{3}{c}{\textbf{TPR (\%)} $\uparrow$}
& \multicolumn{1}{c}{\textbf{FPR (\%)} $\downarrow$}
& \multicolumn{1}{c}{\textbf{FPR (\%)} $\downarrow$}\\
\cmidrule(lr){2-4} \cmidrule(lr){5-5}
\cmidrule(lr){6-8} \cmidrule(lr){9-9}
\cmidrule(lr){10-10}
& \lone{} & \ltwo{} & \lthree{} & \lfour{}  
& \lone{} & \ltwo{} & \lthree{} & \lfour{} & \lfive{} \\
\toprule

Pangram 3.0 
& $100.0$ & $100.0$ & $100.0$ & $3.0$ 
& $97.0$ & $99.3$ & $92.6$ & \hltred{$3.1$} & $0.0$\\
\midrule

GPTZero 
& $96.7$ & $93.3$ & $91.0$ & $3.0$ 
& $96.0$ & $95.8$ & $89.4$ & \hltred{$3.4$} & $1.0$ \\
\midrule

LogLikelihood 
& $97.8$ & $96.4$ & $72.4$ & $0.4$
& $46.0$ & $40.9$ & $30.5$ & $0.0$ & $0.0$\\
{{\textcolor{blue}{with paper context}}}
& $98.9$ & $99.6$ & $91.6^{*}$ & $1.6$ 
& $59.8^{*}$ & $50.5^{*}$ & $43.7^{*}$ & $4.2^{*}$ & $0.4$\\
\midrule

Fast-DetectGPT 
& $100.0$ & $100.0$ & $97.5$ & $3.5$ 
& $72.1$ & $68.2$ & $63.1$ & $4.6$ & $0.2$\\
{{\textcolor{blue}{with paper context}}}
& $100.0$ & $100.0$ & $99.3^{*}$ & $9.0^{*}$ 
& $75.3^{*}$ & $73.9^{*}$ & $68.2^{*}$ & $9.5^{*}$ & $0.9$\\
\midrule

Binoculars 
& $51.1$ & $50.7$ & $39.6^{*}$ & $0.0$ 
& $21.6$ & $22.3$ & $14.1$ & $0.1$ & $0.0$\\
{{\textcolor{blue}{with paper context}}}
& $50.7$ & $53.3$ & $33.7$ & $0.0$ 
& $20.5$ & $20.0$ & $14.3$ & $0.2$ & $0.0$\\

\bottomrule
\end{tabular}
\end{table*}

Further, \textbf{Pangram and GPTZero classify a non-trivial fraction of human-written but AI-polished (\lfour{}) reviews as AI-generated.} 
One strategy to use the predicted labels from Pangram and GPTZero (``AI,'' ``Mixed,'' or ``Human'') for enforcing Polishing-only policies is to penalize only the reviews which are classified as ``AI''. Treating ``Mixed'' in the same category as ``Human'' ensures that reviewers are given maximum benefit of doubt and the detectors are optimized for low FPR on policy-compliant reviews. We use this strategy to report TPRs and FPRs for Pangram and GPTZero in Table~\ref{tab:detector_scores_on_test_data}. Even under this favorable setup, an unacceptable number of \lfour{} reviews are penalized: $3.1\%$ and $3.4\%$,  in \hardsplit{} (see Figure~\ref{fig:conf-mat}). 
Breaking this down by the LLM used for polishing, 
Pangram's FPR on \lfour{} reviews is $0.5\%$ (GPT-5), $5.3\%$ (Gemini-2.5-pro), $1.3\%$ (Gemma-3-27b-it), $3.2\%$ (Llama-3.1-70B), and $5.5\%$ (Qwen-3-30B). 
For GPTZero, the corresponding rates are $1.3\%$ (GPT-5), $2.9\%$ (Gemini-2.5-pro), $1.8\%$ (Gemma-3-27b-it), $1.8\%$ (Llama-3.1-70B), and $9.2\%$ (Qwen-3-30B).
Some examples of these misclassifications are provided in 
Appendix~\ref{appx:misclassifications}. 
On the other hand, under a No-LLM-use policy, reviews classified as ``Mixed'' should also be penalized. Because \lfour{} reviews are prohibited under this policy, enforcing it with Pangram and GPTZero 
would mean that $74.7\%$ and $10.5\%$ of \lfour{} reviews will go undetected, since they are classified as ``Human.''

The  
false positive rates 
on \lfour{} reviews across 
all detectors point to a fundamental limitation 
of current detection systems.
If conference organizers
were to enforce 
Polishing-only policies 
with these detectors, 
an unacceptably large fraction of policy-compliant reviews would be flagged and penalized. Therefore, 
\textbf{``Polishing-only'' policies are not   currently enforceable}. 
\hypertarget{contamination-note}{We note that \textbf{these conclusions are robust to potential data contamination}}. 
It is conceivable 
that the training data of the LLMs used to generate reviews or the AI text detectors may have included the original human reviews from PeerRead dataset.
Firstly, this may enable LLMs to produce more human-like reviews making our evaluation dataset more challenging. Since some detectors (the proprietary ones) are able to detect purely AI-generated reviews regardless, this scenario does not weaken our conclusions. Secondly, if detectors like Pangram and GPTZero saw these reviews during their training on web-scale human-AI parallel data, our results constitute an optimistic estimate of their performance in the wild, despite which they fail to meet the accuracy standards required for enforcing Polishing-only policies. Therefore, potential contamination does not affect our conclusions.
A more subtle concern is that possible contamination could cause our evaluation to underestimate Pangram's false positive rate on fully human-written reviews. However, Pangram's publicly available technical documentation independently reports a false positive rate of approximately 1 in 10,000~\citep{emi2025falsepositives} on fully human-written content. If this self-reported FPR is to be trusted, then the $0\%$ FPR we observe on human reviews is unlikely to be an artifact of contamination.

\begin{figure}[t]
    \centering
    \includegraphics[width=0.95\textwidth]{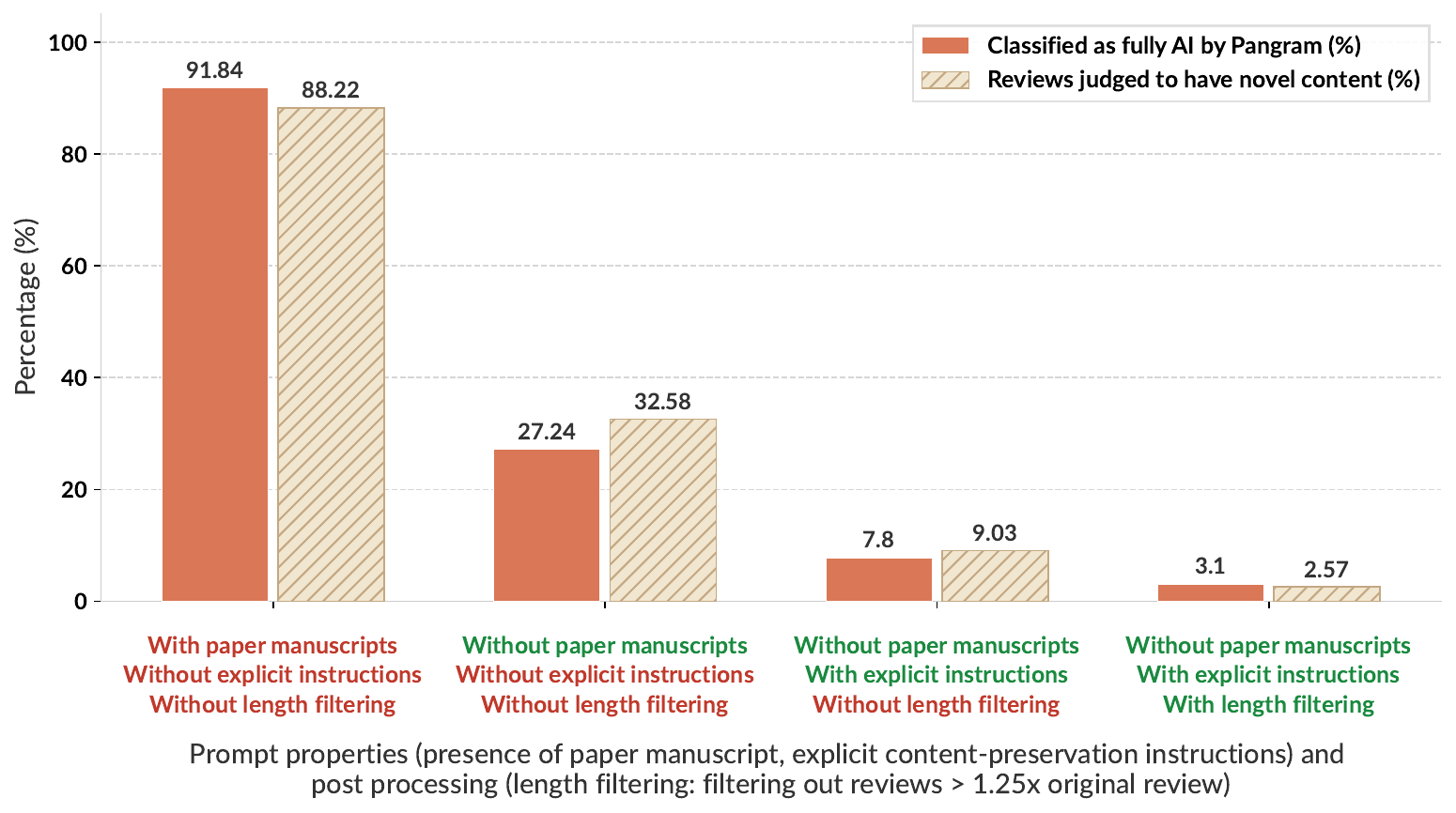}
    \caption{
    \textbf{LLMs inadvertently introduce new content rather than merely ``polishing'':}
    Including paper manuscript, omitting \textit{explicit content-preservation instruction} and specifying generous word limits in the prompt leads to LLMs introducing new content in the ``polished'' review. 
    Interestingly, such reviews are more likely to be flagged as AI-generated by Pangram.
    }
    \label{fig:polishing-only-prompt}
\end{figure}

Surprisingly, we observe that
\textbf{LLMs occasionally introduce new content
to the human review 
when instructed to polish it 
for improving grammar and clarity}.
We observe that three factors
pertaining to the structure of the polishing-only prompt
lead to such behaviour: 
(1) attaching the paper manuscript, (2) omitting \textit{explicit instruction} to preserve technical content, and (3) specifying a word limit that is higher than the length of the human-written draft review. 
In the third case, LLMs tend to treat the word limit as the target length as opposed to an upper bound and generate longer polished reviews. 
We examine these effects systematically at scale 
by employing LLM-as-a-judge 
to flag if a polished review introduces new content relative to the original human review (more details in Appendix~\ref{appdx:verify-new-content}).
We study four polishing strategies
where the generation is subject to increasingly stricter safeguards---starting with prompts featuring all the three above-mentioned properties and progressively removing one at a time.
For these strategies, Figure~\ref{fig:polishing-only-prompt} reports 
the percentage of polished reviews 
judged to include new content, 
alongside the fraction of reviews classified as ``AI'' by Pangram.
Clearly, both these percentages are highly correlated.
This correlation motivates our choice of 
prompts (explicit content preservation instructions,
no access to paper manuscript) and 
post-processing (filtering out reviews $1.25$ times the original length) 
for constructing our \lfour{} reviews
to limit the LLM's role to polishing.
It is important to note that 
all \lfour{} results reported
elsewhere in the paper
use reviews generated following these safeguards.
For the $\sim 3\%$ of these reviews flagged as fully AI-generated
by commercial detectors (Table \ref{tab:detector_scores_on_test_data}), 
we manually verify that the misclassifications are not due to new technical content 
and are genuine false positives.
More importantly, from a practical perspective, 
these findings provide actionable insights for reviewers seeking to use AI for polishing while avoiding false accusations, \textbf{reviewers
should phrase their prompts clearly, with explicit instructions 
to not include new content, 
and should verify the resulting outputs carefully}.

A broad spectrum of reviews discussed here,
while not strictly polishing-only,
are more accurately characterized as mixed human-AI authorship
rather than fully AI generated.
This includes reviews where the reviewer's intent was to comply 
with the polishing-only policy,
but the LLM inadvertently introduced new content 
(e.g., due to omitting explicit content-preservation instructions 
or specifying a generous word limit), 
as well as reviews involving more substantive AI assistance, 
such as those generated with access to the paper manuscript or from human-provided key points (\lthree{}).
In all these cases, the reviews retain significant human-authored input. 
Yet, Pangram overwhelmingly classifies them as ``AI'' rather than ``Mixed'', around $20\%$ for the former category and over $90\%$ for the latter.
This distinction matters when interpreting estimates of AI-generated reviews.

Pangram Labs recently used their AI-detection tools to analyze ICLR 2026 reviews and submissions, claiming that 21\% of reviews were fully AI-generated~\citep{pangram21percentage} and flagging many individual reviews and submitted papers as ``fully-AI generated''~\citep{ICLRPangram2026}. However, our findings suggest that \textbf{both these claims likely misrepresent the extent of fully AI-generated content}: the former may include a substantial fraction of mixed-authorship reviews, and a non-trivial number of the latter might be misclassifications given the scale of ICLR.

\subsection{What is the impact of ``humanization'' (adversarial paraphrasing)?}

In our earlier experiments, the detectors already struggle with
the \hardsplit{},
even without humanization. We therefore focus the analysis on the \basesplit{}, where baseline performance is sufficiently strong to reveal the impact of humanization. These results are summarized in Table~\ref{tab:hum_detector_scores}.

\begin{table}
\caption{
Post humanization, open-source detectors classify all reviews as human. Similarly, Pangram and GPTZero suffer notable drops in their ability to detect AI-generated reviews if they are humanized. Importantly, \lfour{} reviews are more frequently detected as AI by Pangram and GPTZero after humanization.
}
\label{tab:hum_detector_scores}
\centering
\small
\setlength{\tabcolsep}{4pt}

\begin{tabular}{ccccc}
\toprule
& \multicolumn{2}{c}{\textbf{\lone{} TPR (\%)} $\uparrow$} 
& \multicolumn{2}{c}{\textbf{\lfour{} FPR (\%)} $\downarrow$} \\
\cmidrule(lr){2-3} \cmidrule(lr){4-5}
\textbf{Detector} 
& \shortstack{\textbf{pre} \\humanization}
& \shortstack{\textbf{post} \\humanization}
& \shortstack{\textbf{pre} \\humanization}
& \shortstack{\textbf{post} \\humanization} \\
\midrule

Pangram & $100$ & $44.6$ & $3.0$ & $3.63$ \\
GPTZero & $96.7$ & $86.7$ & $3.0$ & $65.6$ \\
LogLikelihood & $96.7$ & $0.0$ & $0.0$ & $0.0$ \\
FastDetect & $100$ & $0.0$ & $2.6$ & $0.0$ \\
Binocular & $51.1$ & $0.0$ & $0.0$ & $0.0$ \\
\bottomrule
\end{tabular}

\end{table}

We find that \textbf{detectors are impaired by humanization of both fully AI-generated and AI-polished reviews.} 
Pangram and GPTZero see a significant drop in their ability to detect fully AI-generated reviews after humanization. More surprisingly, humanization results in higher false positives for human written but AI polished reviews relative to the pre-humanization setting, marginally for Pangram but dramatically for GPTZero. FastDetect, Binoculars and LogLikelihood, on the contrary, classify all reviews as human-written post humanization (both AI and \lfour{}) (Table~\ref{tab:hum_detector_scores}).

\textbf{Both Pangram and GPTZero can deal with humanization under No-LLM-use policy.} Pangram classifies $44.6\%$ of humanized \lone{} reviews as ``AI", and an additional $48.2\%$ as ``Mixed". Therefore, under a No-LLM-use policy, where ``Mixed" reviews are also unacceptable, Pangram successfully flags $92.8\%$ of humanized policy violations. 
GPTZero, on the other hand, classifies  $86.7\%$ of humanized \lone{} reviews as ``AI" but none as ``Mixed''. 
Hence, unlike Pangram, a blanket ban on LLM-assistance will not help its ability to detect violations. 
However, its humanized \lfour{} FPR of $65.6\%$ renders it unsuitable for enforcing a Polishing-only policy.
This makes No-LLM-use policy more enforceable in presence of humanization tools.

\subsection{What additional context does the peer review setting provide, and is that useful for AI detection?}
\label{subsec:exps-pr-specific-adv}

We identify two advantages of the peer review setting that can be potentially leveraged to improve detection performance: (1) availability of the manuscript of the paper under review, and (2) restriction to a specific writing style and the relatively narrow domain of scientific peer reviews. We begin with the former examining how access to the manuscript itself can be leveraged for detection. The paper PDF is often the majority of the prompt that goes into an LLM to generate a review. It is, therefore, valuable additional context. We explore the following two strategies to make use of it:

\begin{wrapfigure}{t}{0.52\textwidth}
	\centering
	\includegraphics[width=0.5\textwidth]{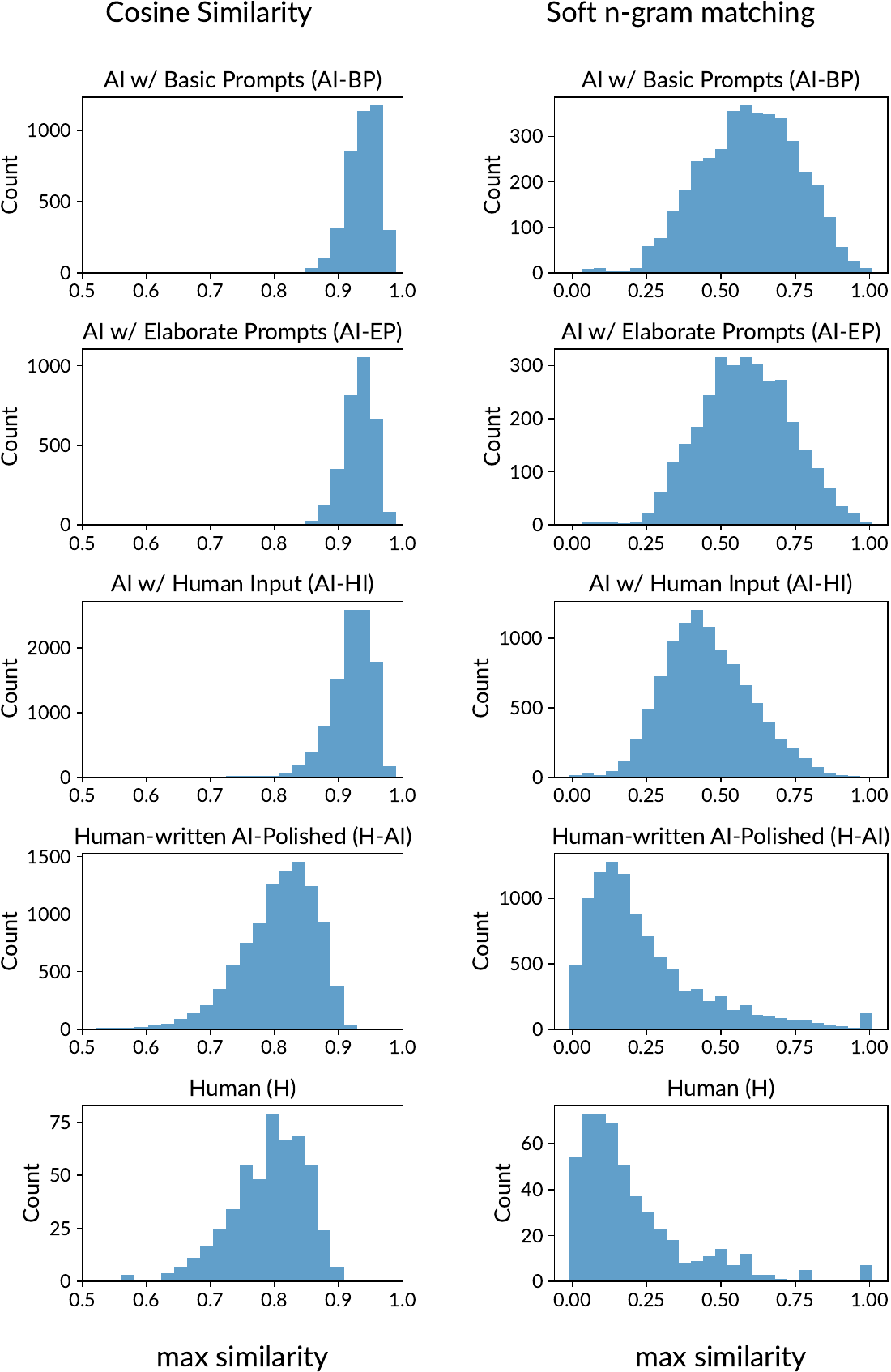}
	\caption{
    \centering 
    Distribution of similarity (maximum over all AI-generated references) across levels of AI assistance. Many individual AI-polished reviews receive similarity scores indistinguishable from those of \fullyAI{} reviews.  
    }
	\label{fig:anchor-similarity-distri}
\end{wrapfigure}

\textbf{(a) Conditioning likelihood on the paper contents.} 
Zero-shot detectors rely on probability of the candidate text under a scoring language model. 
LogLikelihood, Fast-DetectGPT and Binoculars fall under this class of detectors.
We can augment these detectors with paper context by simply computing this probability conditioned on the contents of the paper.
We choose the Abstract, Introduction and Conclusion of the paper as relevant context since some scoring models cannot fit the entire paper in their context window. The results are summarized in Table~\ref{tab:detector_scores_on_test_data} 
alongside that of their regular context-agnostic variant.
Incorporating paper context yields statistically significant improvements in detection rates of AI$^*$ for LogLikelihood and Fast-DetectGPT. 
However, even with these gains, true positive rates remain below $80\%$ on the \hardsplit{}, with significantly higher false positives 
on H-AI reviews.
Therefore, we conclude that
\textbf{the additional context of paper manuscript
yields measurable gains 
in the performance of zero-shot detectors,
albeit not to an extent 
that makes them usable.}

\textbf{(b) Similarity with AI-generated reference reviews.}\label{para:reference-ai-reviews} In this approach, we first generate several AI-generated reviews by prompting LLMs with the paper manuscript. We hypothesize that higher similarity with these references is an indicator of greater AI involvement. This method has been previously shown to be more effective than zero-shot detectors at detecting fully AI-generated peer reviews~\citep{yu2025your}. Related work has also explored similarity-based measures for assessing AI involvement in text. EditLens~\citep{thai2025editlens} quantifies AI editing by estimating similarity to a hypothetical human-written version of the same text. Our approach is conceptually similar, but avoids reliance on such a hypothetical human reference and instead leverages multiple AI-generated references.

\textbf{Similarity metrics:} 
We start with the cosine similarity between the pretrained embeddings of the candidate and the reference review~\citep{yu2025your}. Additionally, towards a more granular measure, we decompose the candidate and reference review into smaller semantic units and measure what fraction of these units in the candidate have a close match in the reference. The similarity metric is called \textbf{Soft n-gram matching} if the semantic units are n-grams~\citep{thai2025editlens}, and \textbf{Soft keypoint matching} if the semantic units are key ideas extracted from the text (implementation details in Appendix~\ref{appdx:ref-ai-revs-others}).

For every paper, we use $3$ 
    models (GPT-5, Gemma-3-27b-it, Qwen-3-30B-thinking), $3$ prompts and $5$ roll outs per model-prompt pair to generate 
    $45$
    references.
We find that \textbf{predictions about individual reviews, based on similarity to AI-generated references, are unreliable.}
We observe a consistent monotonic shift in the distribution of similarities towards lower values as we move from completely AI-generated (\lone{}) to completely human-written (H) reviews (Figure~\ref{fig:anchor-similarity-distri}).
However, these distributions overlap substantially enough that individual reviews cannot be reliably classified.
To further illustrate this limitation, we train a five-class XGBoost~\citep{chenxgboost} classifier 
using similarity scores with all $45$ references 
as the feature vector, sorted to 
ensure permutation invariance 
across reference orderings. 
At inference, examples predicted as \lone{}, \ltwo{} or \lthree{} are mapped to the ``positive'' (AI-generated) class.\footnote{\label{fn:positiveclarify}The definitions of TPR and FPR are adjusted accordingly (e.g. a \ltwo{} review classified as ``AI'' is a true positive, whereas a \lfour{} review classified as ``AI'' is a false positive)} 
With this, more than $4\%$ of AI-polished reviews are misclassified as AI-generated (Table~\ref{tab:similarity_methods}) with each of the similarity metrics. 
Additionally, and concerningly, we see non-zero FPR on fully human-written (\lfive{}) reviews.
This confirms the unreliability of similarity-based detection.
We speculate that pretrained embeddings may not adequately capture the differences in reasoning, style and argumentation that distinguish human writing from AI.
\begin{table}[t]
\caption{\centering Classifier trained on similarity to AI generated references as features. High error rates across-the-board indicate infeasibility of  individual review level predictions with these features.}
\label{tab:similarity_methods}

\centering
\small
\setlength{\tabcolsep}{6pt}
\begin{tabular}{lccccc}
\toprule
& \multicolumn{3}{c}{\textbf{TPR (\%)} $\uparrow$} & \multicolumn{2}{c}{\textbf{FPR (\%)} $\downarrow$} \\
\cmidrule(lr){2-4} \cmidrule(lr){5-6}
\textbf{Similarity metric} & \lone{} & \ltwo{} & \lthree{} & \lfour{} & H \\
\toprule

Cosine similarity & $98.5$ & $98.2$ & $90.8$ & \hltred{$8.2$} & \hltred{$4.0$} \\

\midrule

Soft n-gram match & $98.7$ & $98.5$ & $93.8$ & \hltred{$4.0$} & \hltred{$2.0$} \\

\midrule

Soft keypoint match & $98.8$ & $97.7$ & $88.9$ & \hltred{$4.4$} & \hltred{$3.0$} \\
\bottomrule
\end{tabular}

\end{table}

These observations highlight limitations of the strategies leveraging the paper manuscript, and motivate us to
turn to the second peer-review-specific advantage: the restricted domain of scientific reviews.
Supervised classifiers for AI text detection are known to excel on in-domain data but struggle with text drawn from a wider range of domains~\citep{bakhtin2019realfakelearningdiscriminate, uchendu-etal-2020-authorship, pu2023deepfake}. However, peer reviews, and academic discourse in general,
follow a 
certain
writing style and constitute a 
relatively 
narrow
domain. 
Availability of review data across different levels of AI-assistance makes supervised training a strong candidate to leverage these stylistic and domain regularities. We explore two sets of features for training classifiers.

\textbf{(a) Stylometric features.} Stylometric and linguistic features have been previously used for authorship attribution and AI-text detection tasks~\citep{ling-feat-1, ling-feat-2, ling-feat-3}. We choose a set of $38$ such features (exhaustively listed in Appendix~\ref{appdx:linguistic_feature_definitions}) capturing lexical diversity (e.g. type-token ratio, bigram, trigram uniqueness, etc.), POS tags (e.g. verb percentage, abstract noun percentage, etc.), readability (e.g. average syllables per word, Flesch reading ease, etc.) and other statistics.

\textbf{(b) Transformer-based representations.} Supervised classifiers built on pretrained transformer representations have been found 
to be
effective for AI-generated text detection, particularly in narrow domains~\citep{zellers,loglikelihood,rodriguez-etal-2022-cross,semevaltask8dlmethods2024}. We do a full finetuning of a RoBERTa model~\citep{roberta2019} with a $5$-label classification head. We train the classifier on overlapping segments of fixed length from review texts. Final prediction is obtained through majority voting over segment-level predictions.

We first report performance in the in-distribution setting where training and test sets are disjoint subsets of the same split (\base{} or \hard{}) in Table~\ref{tab:supervised-detectors-id}. 
We conduct further experiments to simulate a more realistic scenario where the classifier encounters reviews generated by unfamiliar models during deployment.
Table~\ref{tab:supervised-detectors-ood} presents these results, where each row corresponds to holding out one model entirely from the training data while training on reviews from all other models.
For both the classifiers described above, following earlier discussion, we treat reviews classified as \lone{}, \ltwo{} and \lthree{} as ``positive'' or AI-generated.\footref{fn:positiveclarify}

\begin{table*}[t]
\captionof{table}{
Evaluating supervised classifiers on examples generated with same models and prompts seen during training represents an optimistic in-distribution setting. Under this setting, full finetuning of a RoBERTa classifier yields better performance than off-the-shelf detectors, including proprietary ones.
}
\label{tab:supervised-detectors-id}
\centering
\vspace{0pt}
\small
\setlength{\tabcolsep}{3pt} %

\begin{tabular}{lcccccccccc}
\toprule
& \multicolumn{5}{c}{\textbf{\Basesplit{}}}
& \multicolumn{5}{c}{\textbf{\Hardsplit{}}}
\\
\cmidrule(lr){2-6} \cmidrule(lr){7-11} 
\textbf{Classifier}
& \multicolumn{3}{c}{\textbf{TPR (\%)} $\uparrow$}
& \multicolumn{2}{c}{\textbf{FPR (\%)} $\downarrow$}
& \multicolumn{3}{c}{\textbf{TPR (\%)} $\uparrow$}
& \multicolumn{2}{c}{\textbf{FPR (\%)} $\downarrow$}\\
\cmidrule(lr){2-4} \cmidrule(lr){5-6}
\cmidrule(lr){7-9} \cmidrule(lr){10-11}
& \lone{} & \ltwo{} & \lthree{} & \lfour{} & \lfive{} 
& \lone{} & \ltwo{} & \lthree{} & \lfour{} & \lfive{} \\
\toprule

Stylometric & $99.3$ & $99.6$ & $98.6$ & $1.3$ & $1.1$ 
& $99.1$ & $97.3$ & $95.6$ & $4.3$ & $0.0$\\

RoBERTa-base & $100.0$ & $100.0$ & $100.0$ & $0.9$ & $0.0$ 
& $100.0$ & $100.0$ & $99.9$ & $1.4$ & $0.0$\\

\bottomrule
\end{tabular}
\end{table*}

\begin{table}
\centering
\small
\setlength{\tabcolsep}{5pt}
\caption{Performance of supervised classifiers on reviews generated by a model held out during training. Under this setting, supervised classifiers can exhibit two failure modes on reviews from the held-out model: they either fail to detect AI$^*$ reviews or produce non-trivially high false positive rates on \lfour{} reviews, limiting their utility in scenarios where reviewers can employ new models and prompts not seen during training. 
\textcolor{darkred}{\textbf{Dark red}} represents \textit{some} selected concerning values.
}
\label{tab:supervised-detectors-ood}
\begin{tabular}{cccccccccccc}
\toprule

&
& \multicolumn{5}{c}{\textbf{Stylometric}}
& \multicolumn{5}{c}{\textbf{RoBERTa-base}}\\
\cmidrule(lr){3-7} \cmidrule(lr){8-12}

\multicolumn{2}{c}{\multirow{2}{*}{\makecell{\textbf{Holdout}\\\textbf{model}}}}
& \multicolumn{3}{c}{\textbf{TPR (\%)} $\uparrow$}
& \multicolumn{2}{c}{\textbf{FPR (\%)} $\downarrow$}
& \multicolumn{3}{c}{\textbf{TPR (\%)} $\uparrow$}
& \multicolumn{2}{c}{\textbf{FPR (\%)} $\downarrow$}\\

\cmidrule(lr){3-5} \cmidrule(lr){6-7}
\cmidrule(lr){8-10} \cmidrule(lr){11-12}

 & 
& \lone{} & \ltwo{} & \lthree{} & \lfour{} & \lfive{}
& \lone{} & \ltwo{} & \lthree{} & \lfour{} & \lfive{} \\
\toprule

\multicolumn{2}{c}{GPT-5}
& $98.7$ & $97.5$ & $93.0$ & \hltred{$3.6$} & $0.0$  &
$100.0$ & $100.0$ & $100.0$ & $1.6$ & $0.0$ \\

\multicolumn{2}{c}{Gemini-2.5-pro}
& $90.7$ & \hltred{$85.8$} & $76.2$ & \hltred{$3.7$} & \hltred{$1.0$} & 
$100.0$ & $100.0$ & $96.3$ & \hltred{$4.6$} & $0.0$ \\

\multicolumn{2}{c}{Gemma-3-27b-it}
& $99.3$ & $95.8$ & $93.8$ & \hltred{$3.5$} & \hltred{$1.0$} & 
$100.0$	& $100.0$ & $95.5$ & $0.0$ & $0.0$ \\

\multicolumn{2}{c}{Llama-3.1-70B-instr}
& \hltred{$45.3$} & $39.2$ & $52.2$ & $1.3$ & $0.0$ & 
$92.0$	& \hltred{$87.5$} & $86.3$ & $0.3$ & $0.0$ \\

\multicolumn{2}{c}{Qwen-3-30B-thinking}
& \hltred{$88.0$} & $85.0$ & $81.8$ & \hltred{$11.0$} & $0.0$ & 
$100.0$	& $100.0$ & $100.0$ & \hltred{$11.5$} & $0.0$ \\

\bottomrule
\end{tabular}
\end{table}

\textbf{In the in-distribution setting, supervised classifiers trained on the review dataset surpass the performance of off-the-shelf detectors}, including proprietary ones (Table~\ref{tab:supervised-detectors-id}). 
The stylometric classifier achieves $95.6\%$ detection of AI-HI reviews but a $4.3\%$ false positive rate on H-AI reviews. The RoBERTa classifier improves upon this, reducing false positives on H-AI reviews to just $1.4\%$ while detecting $99.9\%$ of AI-HI reviews,
outperforming the proprietary detectors.
However, \textbf{when evaluated on reviews from models held out during training, both classifiers exhibit inconsistent generalization across held out models} (Table~\ref{tab:supervised-detectors-ood}). 
For instance, when Llama-3.1-70B is held out, the RoBERTa classifier's ability to detect \fullyAI{} reviews drops substantially (as low as $87.5\%$ on \ltwo{}). 
For others, such as Qwen-3-30B, the classifier maintains near-perfect detection of \fullyAI{} reviews, 
but at the cost of an $11.5\%$ false positive rate on \lfour{} reviews. 
While the classifier generalizes well to certain held-out models (e.g., GPT-5 and Gemma-3), 
conference organizers cannot know in advance which models reviewers will use 
given the frequency of new LLM releases.
Hence the fact that the performance can degrade unpredictably depending on which model is absent in training data is itself a limitation.
The stylometric classifier demonstrates similar model-dependent failures.
This implies that 
in order to use supervised detectors for enforcing policies in the wild, 
\textbf{their training has to keep up with the rapid pace of new model releases}; otherwise, reviewers will routinely employ systems unseen during detector training.

\section{Conclusion}

In this work, we examined the feasibility of enforcing policies that permit LLM use for polishing peer reviews while prohibiting more substantial AI assistance by evaluating the capabilities and limitations of current AI text detectors.
We curated a dataset of more $50,000$ LLM-generated, LLM-assisted and purely human-written paper reviews, and  
our analysis of popular text detectors on these reviews revealed that detectors 
misclassify a non-trivial fraction AI-polished reviews as AI generated.  Our findings indicate that Polishing-only policies
cannot be reliably enforced with current detection technology, 
as detectors are unable to consistently distinguish 
policy-compliant AI-polished reviews from 
policy-violating AI-generated ones.

Our work also suggests that broader claims
based on estimates from AI detectors in academic outlets
and social media discussions are to be interpreted 
with caution. 
For instance, recent claims by Pangram Labs that 
$21\%$ of ICLR 2026 reviews were fully AI-generated, as well as tagging each individual paper and review in terms of AI content(~\citealp{pangram21percentage},~\citealp{ICLRPangram2026}), 
may conflate mixed-authorship reviews with fully AI-generated ones, 
potentially overstating the extent of policy violations.

Our evaluation is a step towards better transparency about false positive rates, 
and indirectly impacts other areas that 
see use of AI text detectors, 
for instance, detecting AI use in student assignments.
We also acknowledge that greater transparency 
comes with a potential downside. 
Public awareness of high false positive rates may enable non-compliant reviewers to claim innocence on grounds of detector unreliability, making it harder to identify actual violations and worsening the enforcement of polishing-only policies.

Finally, we also explored several strategies to enhance existing open-source detectors by leveraging advantages unique to the peer-review setting, such as manuscript context, domain constraints, and similarity to AI-generated reference reviews. However, none of these approaches yielded broadly reliable fine-grained detection. That said, these signals did yield measurable improvements in certain settings, and future work could improve upon them. Such improvements may have implications beyond peer review to other scenarios where additional contextual information about the generation process is available.

\section*{Acknowledgments}
The work of NBS was supported by NSF 1942124, 2200410 and ONR N000142512346. A subset of Pangram API requests were supported through the Pangram Academic Plan. 
Findings, conclusions and recommendations expressed here are those of the authors alone and do not necessarily reflect the views of our sponsors.

\bibliography{bibtex}
\bibliographystyle{plainnat}

\newpage
\appendix

\appendix
\noindent {\bf \Large Appendices}\\

\setcounter{tocdepth}{2}
\titlecontents{section}[1.5em]{\addvspace{2em}\bfseries}
  {\contentslabel{1.5em}}{\hspace*{-1.5em}}
  {\titlerule*[0.6pc]{.}\contentspage}
\titlecontents{subsection}[3.8em]{\addvspace{1em}}
  {\contentslabel{2.3em}}{\hspace*{-2.3em}}
  {\titlerule*[0.6pc]{.}\contentspage}
\startcontents[appendix]
\printcontents[appendix]{}{1}{\subsection*{Appendix Contents}}
\vspace{1em}

\section{Verifying Extent of New Content in LLM-Polished Reviews}
\label{appdx:verify-new-content}

We pair source human reviews (Review A) 
with their LLM-polished counterparts (Review B) 
and employ Gemini as a judge to evaluate each pair on the question: \textit{Does Review B introduce any new technical information that is not present in Review A?}
Since summaries may be expanded 
for comprehensiveness rather than substance, 
we ignore any new information introduced there 
and consider only sections requiring substantive human judgement.  
Figure~\ref{fig:polishing-only-prompt} reports the percentage of reviews 
where the judge answered affirmatively to the above question. 
To assess whether Gemini can serve as a reliable judge, 
we perform two complementary validation checks. 

\begin{table}[t]
\centering
\small
\setlength{\tabcolsep}{4pt}
\caption{\centering
Inter-annotator agreement and 
Gemini-as-a-judge accuracy 
on whether LLM-polished reviews
introduce novel content relative to 
their source human reviews, 
with two independent human annotators.
}
\label{tab:new_information_check}
\begin{tabular}{cccc}
\toprule
\multicolumn{3}{c}{\textbf{Cohen's $\kappa$ (95\% CI)}} &
\multicolumn{1}{c}{\textbf{Gold standard annotations}} \\
\cmidrule(lr){1-3}
\cmidrule(lr){4-4}
Human A~-~Human B &
LLM~-~Human A &
LLM~-~Human B &
Gemini Judge Accuracy \\
\midrule

$0.68\;[0.47, 0.88]$ &
$0.76\;[0.57, 0.92]$ &
$0.76\;[0.60, 0.92]$ &
$88.0$ \\
\bottomrule
\end{tabular}
\end{table}

\begin{wraptable}[10]{r}{0.42\textwidth}
\vspace{-1em}
\caption{\centering Interpretation of Cohen’s $\kappa$ agreement~\cite{landiskoch}}
\label{tab:kappa_stats}
\centering
\footnotesize
\setlength{\tabcolsep}{3.5pt}
\begin{tabular}{lc}
\toprule
\textbf{Cohen's $\kappa$} & \textbf{Agreement} \\
\midrule
$\leq 0.00$ & Poor \\
$0.00-0.20$ & Slight \\
$0.21-0.40$ & Fair \\
$0.41-0.60$ & Moderate \\
$0.61-0.80$ & Substantial \\
$0.81-1.00$ & Almost Perfect \\
\bottomrule
\end{tabular}
\end{wraptable}

\paragraph{Inter-annotator agreement.} 
We randomly sample 50 review pairs, 
25 labeled ``yes'' 
and 25 labeled ``no'' 
by the LLM judge, 
and present them to 
two authors of this paper 
as independent human annotators. 
We compute Cohen’s $\kappa$~\citep{cohen1960coefficient}
to measure inter-annotator agreement between the two human annotators, 
as well as the agreement between 
the Gemini judge and each human annotator  
(Table~\ref{tab:new_information_check}).
For both human annotators,
LLM-human agreements 
are ``substantial'' (Table~\ref{tab:kappa_stats},
~\citealp{landiskoch}),
which is consistent with the human-human agreement. 

\paragraph{Gold standard annotations.} 
We construct a gold-standard annotation set for the same $50$ pairs using the human annotations.
When both annotators concur, we adopt that label.
In cases of disagreement, the two annotators assign a consensus label through discussion. 
We subsequently compare the predictions of Gemini against this gold standard and 
find that it achieves $88\%$ accuracy (Table~\ref{tab:new_information_check}).

\paragraph{An example.}
Following example demonstrates a case where the LLM inadvertently adds new information to the polished review in response to a well-meaning polishing-only prompt. We denote such additions with \content{green}. The polishing prompt used to generate this \lfour{} example lacked explicit instruction for content preservation. \struct{Blue} highlighting indicates cases where the content is preserved but the sentence structure or ordering of ideas differs. 

\begin{itemize}
\item \textbf{Human-written review excerpt:} 
... Pros: The few-shot learning results look good, but I'm not an expert in this area. The idea of using a ``double'' variational bound in a hierarchical generative model is well presented and seems widely applicable. Questions: \struct{When training the statistic network, are minibatches used?} ...  \struct{If your statistic network is trained on minibatches, it won't be able to learn this characterization, because a given minibatch will be missing some of the examples from the dataset} ... Suggestions:  Hierarchical forecasting (electricity/sales) could be an interesting and practical use case for this type of model.

\item \textbf{LLM-``Polished'' review excerpt:} 
... \content{Decision: Reject}. The paper presents an interesting idea, but the experimental results do not convincingly demonstrate a significant advance ... Supporting Arguments: The core idea of learning representations of datasets is promising. The paper clearly articulates the motivation and positions the work within the relevant literature ... \content{The results on the synthetic 1D distributions and spatial MNIST} ... \content{The most significant results are presented for the Omniglot and YouTube Faces datasets} ... a more detailed analysis of the performance gains, would be necessary to justify acceptance ... \struct{it's unclear whether the statistic network sees all datapoints within each dataset in a minibatch. If not, this introduces a bias in the gradient estimation.} ... Additional Feedback: \content{The paper would benefit from more detailed ablation studies} ... \content{The choice of the sample mean as the pooling method in the statistic network seems somewhat arbitrary} ... In the related work section, \content{the connections to multiple-instance learning (MIL) could be explored in more detail} ... \content{The notation could be slightly improved for clarity.} ... Questions for the Authors: \struct{1.  During training of the statistic network, does a minibatch contain all data points within each selected dataset, or is the minibatch constructed from individual data points across multiple datasets?} ... \struct{Have they considered any strategies to mitigate this bias, such as using a larger batch size or incorporating a mechanism to ensure that all data points are considered over time}
\end{itemize}

\section{Additional \lfour{} variants}
\label{appdx:addtnl-polishing-variants}
Prompting general-purpose LLMs represents one popular mode of AI-based assistance in polishing review drafts. 
We acknowledge the real life polishing practices with AI-assistance can be more diverse. 
Broadly, we intend H-AI to serve as a blanket category for all forms of AI usage allowed by a polishing-only policy. 
\textbf{Grammarly} naturally fits the polishing-only criterion. Although conferences don't take an explicit stand on \textbf{reviews translated with AI-based tools}, such usage should be permitted under polishing-only policy assuming these tools don’t tamper with the technical content significantly. 
Additionally, we consider the scenario where a reviewer supplies \textbf{personalized stylistic instructions} to the LLM during polishing so that the polished output better reflects their natural writing style. 
We conduct experiments to investigate these forms of AI-based polishing.

\paragraph{Grammarly.} For $100$ randomly sampled human reviews from our dataset we manually accept all suggestions (representing maximal modification) in the Grammarly Proofreader web interface. Pangram classifies 89 of these edited reviews as ``Human'' and the rest as ``Mixed'' (none are classified as ``AI''). This small-scale result demonstrates that under polishing-only policy, Grammarly-style edits on human reviews do not pose the risk of false positives under Pangram, and therefore polishing-only policy may be enforced for Grammarly. A larger-scale study, due to absence of API, requires more human effort and we leave that to future work.

\paragraph{LLM-based Translation.} To the best of our knowledge, no open-source dataset of non-English peer reviews is publicly available. As a proxy, we translate human-written English reviews into German (de), Chinese (zh), and Japanese (ja), and treat each translation as a stand-in for a review originally written in that language. These reviews are then translated back to English. 
Reusing a single translator in both directions risks artificially preserving the original phrasing and biasing the evaluation. Therefore, we use different translators, namely Google Cloud Translate\footnote{\url{https://cloud.google.com/translate}} and DeepL\footnote{\url{https://developers.deepl.com/}}, for the forward and backward directions. 
Table~\ref{tab:translator_pairs} reports that a non-trivial percentage of these reviews are classified as fully ``AI'' generated by Pangram.

\begin{table}[t]
\centering
\small
\caption{Percentage of back translated reviews Pangram classifies as fully ``AI''-generated}
\label{tab:translator_pairs}
\resizebox{0.48\textwidth}{!}{%
\begin{tabular}{llccc}
\toprule
Forward Translator & Backward Translator & de & zh & ja \\
\midrule
Google Translate & DeepL            & 1.9 & 1.1 & 3.4 \\
DeepL            & Google Translate & 1.5 & 4.5 & 1.1 \\
\bottomrule
\end{tabular}
}
\end{table}

\paragraph{Personalized polishing instructions.} 
While asking an LLM to polish their review, reviewers can add stylistic instructions, so that the polished output better reflects their idiosyncratic writing style rather than generic LLM phrasing. 
To simulate this, we generate reviews with existing \lfour{} prompts augmented with additional stylistic instructions (specifically, "use British English", "use short sentences", and "do not use semicolons and em-dashes").
Pangram classifies $2.42\%$ of these reviews as ``AI'' generated.

\paragraph{} While Grammarly-based edits appear safe from Pangram-based flagging, the other two forms of polished reviews yield FPRs high enough to keep our unenforceability claim for polishing-only policies intact. Even if we take $1\%$ (the lowest rate observed across our translation experiments) as a conservative lower bound on the FPR across all H-AI variants, hundreds of policy-compliant reviews would still risk being flagged at the scale of a venue like NeurIPS. We note, however, that real-world polishing behaviour is private and largely unobservable in practice. While we aim to be comprehensive, the variants studied here are simply representative of plausible usage rather than an exhaustive enumeration.

\section{Off-the-Shelf Detectors}
\label{appdx:appendix_detectors}

\subsection{Descriptions}
\label{appdx:auto-detect-descp}

In this section, we detail the detectors used as automatic benchmarks in Table \ref{tab:detector_scores_on_test_data}
We evaluate a mix of commercial closed-source and open-source baselines to provide comprehensive coverage of detection approaches.

\begin{itemize}
  \item \textbf{LogLikelihood}~\citep{loglikelihood} is the simplest detection method where we compute the average log-likelihood of the text under a given language model. Higher likelihoods indicate AI-generated text, while lower likelihoods suggest human-written text. Our LogLikelihood detector uses LLaMA-3-8B for scoring.
  \item \textbf{Binoculars}~\citep{binoculars} is an open-source detector that works by contrasting token log-probabilities from an observer language model with those from a performer model to compute a cross-entropy difference signal. We use Falcon-7B as the observer models and Falcon-7B-Instruct as the performer model.
  \item \textbf{Fast-DetectGPT}~\citep{fastdetectgpt} is an open-source detector that improves upon the standard DetectGPT by replacing expensive perturbation-based detection with efficient likelihood-based scoring. It operates by comparing token-level log-likelihoods from a base language model against alternative token choices from a reference model, computing a normalized discrepancy score across the sequence. We use Falcon-7B for scoring and Falcon-7B-Instruct for sampling. 
  \item \textbf{GPTZero}~\citep{tian2026gptzero} is a commercial closed-source detector that uses a deep neural classifier, trained end-to-end on large corpora of human and LLM-generated text from multiple models. As a proprietary system, specific architectural and training details are not publicly disclosed. We use the GPTZero API dated 03.01.2026. We use argmax decoding to choose the highest probability class among ``ai'', ``mixed'' and ``human'' for the final prediction.
  \item \textbf{Pangram}~\citep{pangram, thai2025editlens} is a proprietary detector that uses a transformer-based neural network trained using negative mining with synthetic mirror prompting with human samples and LLM generated samples that closely match them. In our work, we access the latest Pangram version 3.0 dated 03.01.26, which is reportedly based on their recent work on quantifying AI editing~\citep{thai2025pangram}. We use the labels returned in the ``prediction\_short'' field.
\end{itemize}

\subsection{Pangram and GPTZero: \Basesplit{} results}
\label{appdx:easy-split-conf-mat}
\begin{figure}[ht]
    \centering
    \includegraphics[width=0.8\textwidth]{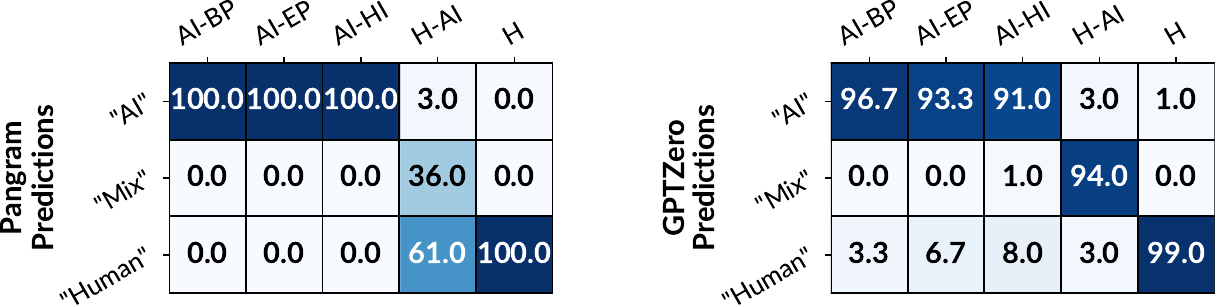}
    \caption{
    Confusion matrices denoting \% of reviews classified as ``AI'', ``Mixed'', ``Human'' by Pangram $\&$ GPTZero on \basesplit{}.
    }
    \label{fig:conf-mat-base}
\end{figure}

Figure~\ref{fig:conf-mat-base} presents the distribution of \basesplit{} reviews over the three outcome categories (``AI,” ``Mixed,” and ``Human”) as predicted by Pangram and GPTZero.

\section{Detection with Peer-Review-Specific Advantages}

\subsection{Context-aware versus Context-agnostic Zero-shot Detection: Statistical Significance Tests}
\label{appdx:statistical_significance}

To test the statistical significance of our results, we use McNemar's test. McNemar’s test is a non-parametric statistical test used to compare whether two related classifiers differ significantly in their predictions on the same set of instances~\citep{McNemar_1947}. The test is based on a $2 \times 2$ contingency table of paired outcomes and considers only the discordant pairs. Let $b$ denote the number of instances misclassified by the first method but correctly classified by the second, and $c$ denote the number of instances correctly classified by the first method but misclassified by the second. The null hypothesis states that neither of the two models performs better than the other.

For sufficiently large numbers of discordant pairs, $(b + c) \ge 25$, we use the continuity corrected McNemar version~\citep {Edwards_1948}, where the McNemar test statistic is computed as,

\[
\chi^2 = \frac{(|b - c| - 1)^2}{b + c},
\]
\[
p = P\!\left(\chi^2_\nu \ge \chi^2\right)
\]

When the total number of discordant pairs is small, $(b + c) < 25$, we use the exact McNemar test, which models $b$ as a binomial random variable under the null hypothesis and compute a two-sided p-value as,
\[
b \sim \mathrm{Binomial}(b + c, 0.5),
\]
\[
p = 2 \cdot \min \left[ P(B \le b),\; P(B \ge b) \right].
\]

In the special case where $b = c = 0$, the p-value is reported as 1, indicating perfect agreement between the paired methods. All McNemar statistics and p-values are computed using the \texttt{statsmodels} Python library~\citep{statsmodel}.

\begin{table}[t]
\centering
\small
\setlength{\tabcolsep}{6pt}
\caption{P-values from McNemar’s test comparing detectors with their context-aware versions.}
\label{tab:mcnemar_results}
\begin{tabular}{lcccc}
\toprule
& \multicolumn{4}{c}{\textbf{p-value}} \\
\cmidrule(lr){2-5}
\textbf{Detector Group} & \lone{} & \ltwo{} & \lthree{} & \lfour{}\\
\midrule
\multicolumn{5}{c}{\textbf{\Basesplit{}}} \\
\midrule
LogLikelihood & $0.37$ & $7.0e^{-3}$ & $1.5e^{-39}$ & $3.0e^{-3}$ \\
FastDetect & $1$ & $0$ & $4.1e^{-4}$ & $3.5e^{-11}$ \\
Binoculars & $1$ & $0.2$ & $5.3e^{-6}$ & $1$ \\
\midrule
\multicolumn{5}{c}{\textbf{\Hardsplit{}}} \\
\midrule
LogLikelihood & $1.6e^{-22}$ & $4.0e^{-13}$ & $2.3e^{-58}$ & $1.7e^{-18}$ \\
FastDetect & $3.1e^{-5}$ & $4.8e^{-7}$ & $9.9e^{-13}$ & $2.8e^{-19}$ \\
Binoculars & $0.11$ & $7.0e^{-3}$ & $0.03$ & $0.62$ \\
\bottomrule
\end{tabular}
\end{table}

\subsection{Detection with Reference AI Reviews: Implementation details}
\label{appdx:ref-ai-revs-others}
In matching-based similarity metrics (Soft n-gram matching and Soft keypoint matching), we consider there is a \textit{close match} between two semantic units (n-grams or keypoints) if the cosine similarity between their vector embeddings is greater than a predefined threshold $\tau$. The final similarity metric is the fraction of semantic units in the candidate review that have a \textit{close match} with at least some semantic unit in any of the references. For Soft n-gram matching, we use $n=40$ (a choice motivated by two competing factors---$n$ should be large enough to capture meaningful semantic information, but not so large that its vector embedding only fits coarse topical information). For Soft keypoint matching, key ideas are extracted using GPT-5-mini with a prompt that asks it to extract important points from the review text. We experiment with the following pretrained embeddings:

\paragraph{Linq-Embed-Mistral.} This is a general-purpose text-embedding model~\citep{linq-embed-mistral} trained for retrieval tasks, which produces $4096$-dimensional embeddings. We load it from the publicly available huggigface checkpoint.\footnote{\url{https://huggingface.co/Linq-AI-Research/Linq-Embed-Mistral}} 

\paragraph{text-embedding-3-small.} We use OpenAI’s general purpose text-embedding model, which produces $1536$-dimensional embeddings, accessed via their API,\footnote{\url{https://api.openai.com/v1/embeddings}} on January~11,~2026.

\paragraph{specter2.} Released by AI2 and designed for scientific text~\citep{allenai2022specter2}, specter2 produces embeddings of size $768$ and we load the publicly available huggingface checkpoint.\footnote{\url{https://huggingface.co/allenai/specter2}}

\paragraph{} Following \citet{thai2025editlens}, the results reported in the main paper (Table~\ref{tab:similarity_methods}, Figure~\ref{fig:anchor-similarity-distri}) are based on Linq-Embed-Mistral as the embedding model. The other embedding models also lead to similar outcomes and the respective numbers are reported in Table~\ref{tab:similarity_methods_appdx}. 

\begin{table}[t]
\caption{\centering Classifier trained on similarity to AI generated references as features. High error rates across-the-board indicate infeasibility of  individual review level predictions with these features.}
\label{tab:similarity_methods_appdx}
\centering
\small
\begin{tabular}{lccccc}
\toprule
& \multicolumn{3}{c}{\textbf{TPR (\%)} $\uparrow$} & \multicolumn{2}{c}{\textbf{FPR (\%)} $\downarrow$} \\
\cmidrule(lr){2-4} \cmidrule(lr){5-6}
\textbf{Similarity metric} & \lone{} & \ltwo{} & \lthree{} & \lfour{} & H \\
\toprule

\multicolumn{6}{c}{\textbf{specter2}} \\

\midrule

Cosine similarity & $93.9$ & $94.3$ & $88.4$ & $12.4$ & $5.0$ \\

Soft n-gram match & $95.7$ & $95.7$ & $89.8$ & $28.0$ & $21.0$ \\

Soft keypoint match & $94.0$ & $90.3$ & $88.5$ & $15.9$ & $11.0$ \\

\midrule

\multicolumn{6}{c}{\textbf{text-embedding-3-small}} \\

\midrule

Cosine similarity & $99.1$ & $98.7$ & $91.9$ & $7.4$ & $2.0$ \\

Soft n-gram match & $96.9$ & $96.8$ & $91.1$ & $6.3$ & $4.0$ \\

Soft keypoint match & $92.8$ & $88.8$ & $80.8$ & $4.9$ & $2.0$ \\

\bottomrule
\end{tabular}
\end{table}

\subsection{Details of Stylometric Features to Characterize Peer Review Writing Style}
\label{appdx:linguistic_feature_definitions}

In this section, we describe the 38 stylometric features used to characterize peer-review writing style and used to train a supervised XGBoost classifier. These features capture lexical diversity, syntactic structure, grammatical composition, sentiment and emotion, and readability.
For part-of-speech tagging, we use the \quotecode{pos\_tag} function of the \quotecode{nltk} python library.\footnote{NLTK: Natural Language Toolkit library (v3.9.2), \url{https://www.nltk.org/}, accessed December 2025.} For syllable-related features, we use the \quotecode{pyphen} python library to hyphenate words and use that as a proxy for syllables.\footnote{Pyphen: Python hyphenation library (v0.17.2), \url{https://pypi.org/project/pyphen/}, accessed December 2025.} For emotion-related features, we define emotion seed words grouped into positive (e.g., good, happy, joy, love), negative (e.g., sadness, anger, fear, guilt), and other emotions (e.g., surprise, empathy). Using pre-trained GloVe 100-dimensional embeddings,\footnote{Pennington et al., \emph{GloVe: Global Vectors for Word Representation}, EMNLP 2014.} words are assigned to emotion categories based on cosine similarity to the corresponding seed words in the embedding space.

\begin{enumerate}

\item \textbf{Average Word Length} - Ratio of number of non-space characters to number of words.

\item \textbf{Average Sentence Length} - Ratio of number of words to number of sentences

\item \textbf{Type Token Ratio (TTR)}~\citep{van2007comparing} - 
TTR is a lexical diversity measured as the ratio of unique words to total words.

\item \textbf{Root Type Token Ratio (RTTR)}~\citep{van2007comparing} - RTTR is a length-normalized lexical diversity. It is computed as $\frac{\text{Unique Words}}{\sqrt{\text{Words}}}$

\item \textbf{Maas Measure}~\citep{tweedie1998variable} 
Maas is a vocabulary richness measure that is robust to text length. It is computed as $\frac{\ln(\text{Words}) - \ln(\text{Unique Words})}{(\ln(\text{Words}))^2}$

\item \textbf{Hapax Legomenon Rate} - 
HLR is the proportion of words that occur exactly once in the text.  
It is computed as $\frac{\text{Words occurring exactly once}}{\text{Words}}$

\item \textbf{Bigram Uniqueness} -
Ratio of unique bigrams to total bigrams.

\item \textbf{Trigram Uniqueness} -
Ratio of unique trigrams to total trigrams. 

\item \textbf{Punctuation Percentage} -
Percentage of characters that are punctuation characters.

\item \textbf{Stop Word Percentage} -
Percentage of words that are stop words. For identifying stop words, we use the English stopwords corpus of the \quotecode{nltk} library.

\item \textbf{Question Percentage} -
Percentage of sentences that are interrogative (end with $?$).

\item \textbf{Exclamation Percentage} -
Percentage of sentences that are exclamatory (end with $!$).

\item \textbf{Abstract Noun Percentage} -
Proportion of total words that are abstract nouns.

\item \textbf{Sparse Abstract Noun Percentage} -
Sparse abstract nouns are low-frequency abstract nouns that are not among the top 5,000 words in the Brown corpus~\citep{francis1964standard}. For this feature, we use the percentage of abstract nouns that are sparse abstract nouns.

\item \textbf{Verb Percentage} -
Percentage of total words that are verbs.

\item \textbf{Sparse Verb Percentage} 
Sparse verbs are low-frequency verbs that are not among the top 5,000 words in the Brown corpus~\citep{francis1964standard}. For this feature, we use the percentage of verbs that are sparse verbs.

\item \textbf{Adjective Percentage} -
Percentage of words that are adjectives.

\item \textbf{Sparse Adjective Percentage} -
Sparse adjectives are low-frequency adjectives that are not among the top 5,000 words in the Brown corpus~\citep{francis1964standard}. For this feature, we use the percentage of adjectives that are sparse adjectives.

\item \textbf{Complex Adjective Percentage} -
We define complex adjectives as those that are morphologically complex (e.g., have suffixes such as \textit{-ive}, \textit{-ous}, \textit{-ic}). For this feature, we use the percentage of adjectives that are sparse adjectives.

\item \textbf{Adverb Percentage} -
Percentage of words that are adverbs. 

\item \textbf{Sparse Adverb Percentage} -
We define sparse adverbs as low-frequency adverbs not among the top 5,000 words in the Brown corpus~\citep{francis1964standard}. For this feature, we use the percentage of adverbs that are sparse adverbs.

\item \textbf{Preposition Percentage} -
Percentage of words that are prepositions. 

\item \textbf{Conjunction Percentage} -
Percentage of words that are conjunctions. 

\item \textbf{Complex Sentence Percentage} -
Percentage of sentences containing at least one subordinating conjunction. 

\item \textbf{Syntax Variety} -
Number of unique POS tags in the text.

\item \textbf{Emotion Word Percentage} -
Percentage of words that are emotion-related. 

\item \textbf{Positive Emotion Word Percentage} -
Percentage of words that are positive-emotion-related. 

\item \textbf{Negative Emotion Word Percentage} -
Percentage of words that are negative-emotion-related. 

\item \textbf{Other Emotion Word Percentage} -
Percentage of words that are other-emotion-related. 

\item \textbf{First-Person Pronoun Percentage} -
Percentage of words that are first-person pronouns. 

\item \textbf{Second-Person Pronoun Percentage} -
Percentage of words that are second-person pronouns. 

\item \textbf{Polarity} - 
We use the \quotecode{textblob} library's sentiment polarity. It uses a Bag-of-Words classifier to obtain the text polarity.\footnote{TextBlob NLP library (v0.19.0), \url{https://textblob.readthedocs.io/}, accessed January 2026.}

\item \textbf{Subjectivity} -
Similar to Polarity, we use the \quotecode{textblob} library's subjectivity score. It uses a Bag-of-Words classifier to obtain the text subjectivity.

\item \textbf{VADER Compound Score}~\citep{hutto2014vader} 
VADER is a lexicon and rule-based sentiment analysis tool. Each word gets a sentiment score, rules are applied according to punctuation, modifiers, negation etc and the final score is normalised. We compute the VADER compound score using the \quotecode{vaderSentiment} python library.\footnote{\texttt{vaderSentiment}: Python implementation of the VADER sentiment analysis model (v3.3.2), available at \url{https://github.com/cjhutto/vaderSentiment}. Accessed January 2026.}

\item \textbf{Average Syllables Per Word} -
Average syllable count per word. 

\item \textbf{Complex Word Percentage} -
Percentage of words with at least three syllables. 

\item \textbf{Flesch Reading Ease}~\citep{flesch1948new} -
Flesch Reading Ease is readability metric based on sentence length and syllable count.
It is computed as $206.835 - 1.015 \times \frac{\text{Words}}{\text{Sentences}} - 84.6 \times \frac{\text{Syllables}}{\text{Words}}$

\item \textbf{Gunning Fog Index}~\citep{gunning1952technique} -
Gunning fog index is a readability metric that estimates the years of formal education required to comprehend a given text. It is computed as $0.4 \times \left(\frac{\text{Words}}{\text{Sentences}} + 100 \times \frac{\text{Complex Words}}{\text{Words}}\right)$

\end{enumerate}

\label{appdx:complete_table_with_stylo_roberta}

\section{Discussion on Policies and Enforcement}
\label{appdx:enforcement-landscape}
\subsection{Current enforcement landscape}
Our main analysis examines the question whether
current AI text detectors reliably distinguish polishing-only-compliant reviews from policy violations? 
Enforcement in practice, however, draws on a broader set of mechanisms. We give a brief overview of the current landscape below.

\paragraph{Trust-based systems and violation reporting.} The predominant and default approach remains trust-based where reviewers are expected to self-certify compliance with their assigned LLM policy and/or disclose their mode of LLM usage~\citep{icml2025reviewerinstructions, iclr2026llmpolicy}. Conferences such as ICML 2025 provide Ethics Violation Reporting forms through which authors and other Program Committee members can report suspected misconduct~\citep{icml2025reviewerinstructions}.

\paragraph{Prompt injection (watermarking).} ICML 2026 deployed a novel enforcement mechanism~\citep{rao2025detecting} based on prompt injection where submission PDFs were watermarked with hidden instructions that, if fed to an LLM, would cause two specific phrases (randomly drawn from a dictionary of ~170,000 phrases) to appear in the generated review~\citep{ICMLblogLLMreview2026}. Reviewers aware of this attack can possibly get around this by either discovering the watermark or editing the review post generation but this scheme targets the most egregious cases of copy-pasted LLM-generated reviews. It detected 795 reviews (~1\% of all reviews) from 506 unique reviewers who had agreed to the no-LLM policy (Policy A), resulting in the desk rejection of 497 papers authored by reciprocal reviewers who violated the policy. Crucially, every flagged instance was also manually verified by a human.

\paragraph{General LLM-text detectors for triage.} ICLR 2026 mentioned having used “LLM detection tools” as a first pass filter to flag reviews for potential violations and escalate them to AC and SACs~\citep{iclr2026llmresponse}. The official blog doesn’t mention the exact tools they used, but GPTZero has independently claimed to have collaborated with ICLR program chairs for reviewing submissions~\citep{gptzero2025iclr}.

\paragraph{Instances of hallucinated references.} Although initial flagging may rely on a combination of ethics violation reporting forms, and AI text detectors used for triage, subsequent action is typically taken only when there is verifiable evidence of LLM misuse, such as hallucinated citations (references to papers, or its authors, that do not exist).

\subsection{Challenges of a No-LLM-use policy}
\label{sec:no-llm-challenges}
 
Our findings on the un-enforceability of polishing-only policies
naturally point toward the No-LLM-use counterfactual. In this
section, we wish to clarify that 
a No-LLM-use policy is not free of challenges either. 
To begin with, a complete prohibition on LLM use
forgoes the productivity gains that responsible AI assistance 
could offer reviewers, especially in the face of rapidly growing submission volumes. 
Beyond this normative concern, the policy also faces
operational challenges in enforcement, which we discuss below.
 
Any detection-based enforcement of a No-LLM-use policy 
will disproportionately catch careless violators, 
while pushing more determined ones toward
stronger evasion strategies. 
This effectively incentivizes bad actors
to behave more adversarially. 
In a No-LLM-use policy, the enforcement task
is to detect whether any form of AI assistance was used 
to write the review.
Our humanization experiments
(Table~\ref{tab:hum_detector_scores}) detectors like Pangram remain reasonably robust
to one-shot automated humanization, flagging $\sim$93\% of humanized AI-BP reviews as either ``AI'' or ``Mixed''. 
However, these experiments only simulate a single pass through an off-the-shelf humanizer. In practice, a determined reviewer can mount more dedicated efforts 
(e.g. repeated iteration through a detector, 
manual rewriting, or human-in-the-loop editing) which are 
plausibly more effective at evading detection than one-shot automated humanization.
We leave a quantitative study of such adversaries to future work.

\section{Human-Written AI-Polished Reviews Flagged as AI}
\label{appx:misclassifications}

In this section, we present excerpts from  
H-AI (Human-written, AI-polished) reviews that were predicted as ``AI'' by Pangram, along with the corresponding sections of their original Human reviews. Despite the high degree of semantic and structural similarity between the two versions, Pangram classifies the H-AI reviews as AI-generated while identifying the Human reviews as human-written. 

We use a highlighting scheme that indicates the nature of the differences between the texts. \vocab{Yellow} highlighting denotes changes in vocabulary choice where the underlying ideas and sentence structure remain unchanged. \struct{Blue} highlighting indicates cases where the content is preserved but the sentence structure or ordering of ideas differs. 

\subsubsection*{Example 1}

\paragraph{Human-written review excerpt:}

This paper provides an \vocab{interesting} analysis of the conditions \struct{which enable generation} of natural looking textures. The \vocab{results} is quite surprising, and analysis is quite \vocab{thorough}.
\vocab{I do think} the evaluation methods \vocab{require more work}, \struct{but as other reviewers mentioned this could be an interesting line of work moving forwards and does not take too much from this current paper which, I think,} should be accepted.

\paragraph{H-AI (Human-written AI-polished) review excerpt:}

This paper presents a \vocab{thorough and insightful} analysis of the conditions \struct{required for generating} natural-looking textures. The \vocab{findings} are quite surprising, and the analysis is \vocab{comprehensive}.

\vocab{While} the evaluation methods \vocab{could be strengthened}, \struct{this limitation does not significantly detract from the paper's core contributions. Further development of the evaluation framework represents a promising direction for future work. Given the novelty of the findings and the quality of the analysis, I believe this paper} should be accepted.

\subsubsection*{Example 2}

\paragraph{Human-written review excerpt:}
In the paper the authors \vocab{try to come up with} a theory which \vocab{works for} graph clustering \vocab{in} model free framework.
However, \struct{the result is} \vocab{naive} and \vocab{unconvincing}. 
\struct{The authors pretend to construct a new theory but theit theory is just a naive handwaving. Just to give a couple of examples:} 
Theorem 1 (Generic Theorem) ... "which also fits G well is close to C" ... Proposition 3 (Informal) ... "under standard recovery conditions" 

\paragraph{H-AI (Human-written AI-polished) review excerpt:}

The authors \vocab{attempt to develop} a theory for graph clustering \vocab{within} a model-free framework. \vocab{Unfortunately, their result falls short}, appearing \vocab{overly simplistic} and unconvincing. \struct{Rather than presenting a genuinely novel theory, the authors seem to be relying on unsubstantiated claims}. For instance, Theorem 1 (Generic Theorem) contains vague assertions, such as "...which also fits G well is close to C". Similarly, Proposition 3 (Informal) lacks rigor, relying on ambiguous "standard recovery conditions".

\section{Prompts Used for Review Generation}
\label{appdx:prompts}

In this section, we provide the full set of prompts used for generating reviews at each of the four levels of our review dataset. Each level contains four to five prompt variants. The Base split has data generated using Prompt Variant 0 at each level, while the \Hardsplit{} has data generated using all prompt variants at each level.

\subsection*{\levelone{} (\lone{})}

\noindent Prompt Variant 0:
\begin{quote}
\small\ttfamily
You are a reviewer at an AI conference. Write a review of the given research paper following the provided reviewer guidelines in 300-400 words. Write only the review. \\
Conference guidelines- \\
\{GUIDELINES\} \\
Paper- \\
\{PAPER\_CONTENT\}
\end{quote}

\noindent Prompt Variant 1:
\begin{quote}
\small\ttfamily
  You are a reviewer for \{CONFERENCE\}. Read the following paper and write a detailed peer review following the reviewing guidelines below in no more than 1000 words. \\
  Conference reviewing guidelines: \\
  \{GUIDELINES\} \\
  Paper: \\
  \{PAPER\_CONTENT\} \\
  Start your response directly with the review text. Do not include any introductory phrases or disclaimers (e.g., 'Sure, here is the review').
\end{quote}

\noindent Prompt Variant 2:
\begin{quote}
\small\ttfamily
 You are a reviewer for \{CONFERENCE\}. In your opinion, how well does the paper meet the reviewing criteria? Write your review with explicit reasoning and justification for your opinions in no more than 1000 words. \\
 Conference reviewing guidelines: \\
  \{GUIDELINES\} \\
  Paper: \\
  \{PAPER\_CONTENT\} \\
  Start your response directly with the review text. Do not include any introductory phrases or disclaimers (e.g., 'Sure, here is the review').
\end{quote}

\noindent Prompt Variant 3:
\begin{quote}
\small\ttfamily
  You are a reviewer for \{CONFERENCE\}. You are an expert in the field relevant to this paper. Provide a detailed and critical peer review, demonstrating deep understanding of the methodology, related work, and implications. Your review should not exceed 1000 words. \\
  Conference reviewing guidelines: \\
  \{GUIDELINES\} \\
  Paper: \\
  \{PAPER\_CONTENT\} \\
  Start your response directly with the review text. Do not include any introductory phrases or disclaimers (e.g., 'Sure, here is the review').
\end{quote}

\noindent Prompt Variant 4: 
\begin{quote}
\small\ttfamily
You are an impartial reviewer for \{CONFERENCE\}. Avoid personal opinions or biases. Base your review purely on objective assessment of clarity, technical soundness, and novelty. Your review should not exceed 1000 words.\\
  Conference reviewing guidelines:\\
  \{GUIDELINES\}\\
  Paper:\\
  \{PAPER\_CONTENT\}\\
  Start your response directly with the review text. Do not include any introductory phrases or disclaimers (e.g., 'Sure, here is the review').
\end{quote}

\subsection*{\leveltwo{} (\ltwo{})}

    For this level, we aggregate the recommended best practices of reviewing from 
	ACL 2017 Last Minute Reviewing Advice,~\footnote{\href{https://acl2017.wordpress.com/2017/02/23/last-minute-reviewing-advice/}{ACL 2017 last minute reviewing advice}}
	NeurIPS 2020 Reviewer Best Practices,~\footnote{\href{https://neurips.cc/Conferences/2020/PaperInformation/ReviewerGuidelines\#:~:text=Reviewer\%20best\%20practices}{NeurIPS 2020 Reviewer Guidelines}}
	NeurIPS 2025 Reviewer Best Practices,~\footnote{\href{https://neurips.cc/Conferences/2025/ReviewerGuidelines\#:~:text=Best\%20Practices,AC\%20right\%20away}{NeurIPS 2025 Reviewer Guidelines}} 
	ICML 2025 Tips for Reviewing,~\footnote{\href{https://icml.cc/Conferences/2025/ReviewerInstructions\#:~:text=and\%20so\%20on.-,Tips\%20for\%20Reviewing,-Before\%20starting\%20to}{ICML 2025 Reviewer Instructions}} 
	and 
	AAAI-26 Guidelines On Writing Helpful Reviews.~\footnote{\href{https://aaai.org/conference/aaai/aaai-26/instructions-for-aaai-26-reviewers/\#:~:text=Guidelines\%20On\%20Writing\%20Helpful\%20Reviews}{Instructions for AAAI-26 Reviewers}} 
	Each prompt variant has a set of best practices generated using one or more of these sources. 

\noindent Prompt Variant 0:
\begin{quote}
\small\ttfamily
  You are a reviewer at an AI conference. Write a review of the given research paper following the provided reviewer guidelines in 300-400 words. Write only the review. Here are some tips and tricks that would help you write a good review. 
  \begin{enumerate}
  \item Identify Claims- Clearly outline the main claims of the paper. Look for key phrases like “The contributions of this paper are. . . ” to identify them. Conference papers usually have 1-2 claims, while journal articles should have several. 
  \item Evaluate Support for Claims- Assess how the claims are supported. Prioritize real-world statistically significant experiments, followed by laboratory experiments, demonstrations, simulations, and theoretical analysis (in decreasing order of reliability). Avoid papers with unexplained data or unsupported claims. 
  \item Assess Usefulness- Determine whether the ideas presented are practically useful. Consider if you or the target audience would use it and why. 
  \item Check Field Knowledge- Ensure the paper reflects common knowledge in the field. Look for correct use of terms and evidence of understanding of relevant literature 
  \item Evaluate Novelty- The work should present a significant improvement or innovation over existing approaches. Ensure references are comprehensive, accessible, and relevant, with proper citations. 
  \item Assess Completeness- Particularly for journal articles, the paper should provide sufficient details for reproducibility. Conference papers may have more limited scope. 
  \item Discuss Limitations-Check if the paper acknowledges its limitations. Journal articles have more space for this, but it is also important for conference papers. 
  \item Be Constructive- Suggest solutions to shortcomings rather than just pointing them out. Focus your criticisms on the paper, not the authors. 
  \item Avoid Bias- Keep your identity anonymous, and ensure your review is impartial and professional. Avoid comments that could indirectly reveal your identity. 
  \item Encourage Potential- If the paper has a good idea but is poorly executed, encourage the authors to revise and resubmit
  \end{enumerate}
  Conference guidelines- \\
  \{GUIDELINES\} \\
  Paper- \\
  \{PAPER\_CONTENT\}
\end{quote}

\noindent Prompt Variant 1:
\begin{quote}
\small\ttfamily
  You are a reviewer for \{CONFERENCE\}. Write a review of the given research paper following the provided reviewer guidelines. Write only the review. Following are some tips that will help you write a good review:
  \begin{enumerate}
    \item Be thoughtful. The paper you are reviewing may have been written by a first year graduate student who is submitting to a conference for the first time and you don't want to crush their spirits.
    \item Be fair. Do not let personal feelings affect your review.
    \item Be useful. A good review is useful to all parties involved: authors, other reviewers and AC/SACs. Try to keep your feedback constructive when possible.
    \item Be specific. Do not make vague statements in your review, as they are unfairly difficult for authors to address. 
    \item Be flexible. The authors may address some points you raised in your review during the discussion period. Make an effort to update your understanding of the paper when new information is presented, and revise your review to reflect this.
    \item Please avoid biasing your review according to discriminatory criteria not having to do with scientific content or clarity. Please avoid wording that may be perceived as rude or offensive. Although the double-blind review process reduces the risk of discrimination, reviews can inadvertently contain subtle discrimination, which should be actively avoided. Example: avoid comments regarding English style or grammar that may be interpreted as implying the author is "foreign" or "non-native". So, instead of "Please have your submission proof-read by a native English speaker,” use a neutral formulation such as "Please have your submission proof-read for English style and grammar issues.”
  \end{enumerate}
  Your review should not exceed 1000 words.  \\
  Conference reviewing guidelines: \\
  \{GUIDELINES\} \\
  Paper: \\
  \{PAPER\_CONTENT\} \\
  Start your response directly with the review text. Do not include any introductory phrases or disclaimers (e.g., 'Sure, here is the review'). 
\end{quote}

\noindent Prompt Variant 2:
\begin{quote}
\small\ttfamily
 You are a reviewer for \{CONFERENCE\}. Write a review of the given research paper following the provided reviewer guidelines. Write only the review. Don’t go over 1000 words. Here is some advice on reviewing:
  Read the paper carefully, critically, and with empathy. 
  After reading the paper, think carefully about whether the paper has properly substantiated the claimed contributions. This may involve verifying proofs, checking whether hypotheses are actually tested by the experiments, checking whether empirical claims do indeed follow from empirical results, etc. Good judgement is needed to determine the severity of any issues that you identify. It is helpful to point out minor issues that are easily fixed, but it is more important to focus on major issues that are critical to the main contributions.
  Consider whether the paper places the research presented into the context of current research. Assessments about a paper’s “originality” and “significance” often crucially depend on how the paper compares to prior works, and thus such prior works should be cited and discussed in the paper.
  Note that in many cases, it is difficult and often unnecessary to cite all related prior works. If some relevant prior works are missed, then think about whether or not including them would change the conclusions of the paper. Some omissions may be considered minor issues that are easily fixed.
  Please give constructive comments and suggestions to the authors to help them potentially improve their paper. In particular, any comments about strengths and weaknesses must be substantiated. \\
  Conference reviewing guidelines: \\
  \{GUIDELINES\} \\
  Paper: \\
  \{PAPER\_CONTENT\} \\
  Start your response directly with the review text. Do not include any introductory phrases or disclaimers (e.g., 'Sure, here is the review').
\end{quote}

\noindent Prompt Variant 3:
\begin{quote}
\small\ttfamily
  You are a reviewer for \{CONFERENCE\}. Write a review of the given research paper following the provided reviewer guidelines. Write only the review. Don’t go over 1000 words. Here are features that make a bad review:
  A review based on a very superficial reading of the paper which asks questions addressed possibly even in the introduction.
  A review based on sentiment and ideology rather than on the merits of the paper. For instance, I do not like LSTMs and thus dislike every paper using them
  A review which claims things about the paper which are unsupported by evidence. Statements like “this has been done previously and is therefore not novel”. Who did it, when, and how?
  Avoid these mistakes. In general, do not use “I”, “you”, “the authors”, etc., in your reviews. Reviews should be depersonalized as much as possible. Use terms like “the paper”, “the work”, “the project”. The review should focus on the work and not the individuals (reviewers or authors). Avoid referring to yourself. If you must refer to yourself, it should be in the third person (e.g., “this reviewer”) and done sparingly. Phrase your comments as would be appropriate if you were speaking respectfully to the authors face-to-face.
  Instead of “What is wrong with this paper?”, ask yourself “How could this paper be better?”
  When suggesting revisions in the review, think about whether the revisions are reasonable in terms of time and resources — which of the recommendations are essential, and which are nice-to-have but optional? \\
  Conference reviewing guidelines: \\
  \{GUIDELINES\} \\
  Paper: \\
  \{PAPER\_CONTENT\} \\
  Start your response directly with the review text. Do not include any introductory phrases or disclaimers (e.g., 'Sure, here is the review')
\end{quote}

\subsection*{\levelthree{} (\lthree{})}

\noindent Key points Generation Prompt:
\begin{quote}
\small\ttfamily
  You are given a review written by a reviewer of an AI conference. You need to give the key points discussed in the review as one or two liners for each point. Write only the key points. \\
  Review- \\
  \{HUMAN\_REVIEW\} \\
\end{quote}

\noindent Prompt Variant 0:
\begin{quote}
\small\ttfamily
  You are a reviewer at an AI conference. You are given the key points to evaluate by your supervisor, you can elaborate on that. Write a review of this research paper following the provided reviewer guidelines in 300-400 words. Write only the review. \\
  Key points- \\
  \{SUMMARIZED\_HUMAN\_REVIEW\} \\
  Conference guidelines- \\
  \{GUIDELINES\} \\
  Paper- \\
  \{PAPER\_CONTENT\}
\end{quote}

\noindent Prompt Variant 1:
\begin{quote}
\small\ttfamily
I am a reviewer for \{CONFERENCE\}. I have read the paper and jotted down the following key points. Help me write the complete reveiw by elaborating on these key points. Don't exceed 1000 words and start your response directly with the review text. Do not include any introductory phrases or disclaimers (e.g., 'Sure, here is the review').\\
  Conference reviewing guidelines:\\
  \{GUIDELINES\}\\
  Paper:\\
  \{PAPER\_CONTENT\}\\
  Key points:\\
  \{SUMMARIZED\_HUMAN\_REVIEW\}
\end{quote}

\noindent Prompt Variant 2:
\begin{quote}
\small\ttfamily
I am a reviewer for \{CONFERENCE\}. I am providing you my initial impression of the paper in the form of some pointers I have noted down. Based on these key points write the complete review following the conference reviewing guidelines.\\
  Conference reviewing guidelines:\\
  \{GUIDELINES\}\\
  Paper:\\
  \{PAPER\_CONTENT\}\\
  Key points:\\
  \{SUMMARIZED\_HUMAN\_REVIEW\}\\
  Start your response directly with the review text. Do not include any introductory phrases or disclaimers (e.g., 'Sure, here is the review'). Please don't exceed 1000 words.
  \end{quote}

\noindent Prompt Variant 3:
\begin{quote}
\small\ttfamily
Help me write a complete review for this \{CONFERENCE\} paper by elaborating on the following points.\\
  Key points:\\
  \{SUMMARIZED\_HUMAN\_REVIEW\}\\
  Conference reviewing guidelines:\\
  \{GUIDELINES\}\\
  Paper:\\
  \{PAPER\_CONTENT\}\\
  Start your response directly with the review text. Do not include any introductory phrases or disclaimers (e.g., 'Sure, here is the review'). Please don't exceed 1000 words.
\end{quote}

\subsection*{Human-written with AI polishing (\lfour{})}

\noindent Prompt Variant 0:
\begin{quote}
\small\ttfamily
I am a reviewer for a renowned scientific conference. Help me paraphrase my review while keeping the original structure, meaning, and technical content intact. Write only the review.  \\
  Review: \\
  \{HUMAN\_REVIEW\} \\
\end{quote}

\noindent Prompt Variant 1:
\begin{quote}
\small\ttfamily
  I am a reviewer for \{CONFERENCE\}. The following review was written by me and I am a non-native English speaker. Your task is to improve the writing for fluency, clarity and natural phrasing, as a native speaker would write it. You must preserve all technical arguments, criticisms and core meaning without alteration. Your job is only to polish the review. The review should not exceed 1000 words. \\  
  Draft review: \\
  \{HUMAN\_REVIEW\} \\
  Start your response directly with the review text. Do not include any introductory phrases or disclaimers (e.g., 'Sure, here is the review').
\end{quote}

\noindent Prompt Variant 2:
\begin{quote}
\small\ttfamily
  I am a reviewer for \{CONFERENCE\}. The following is a draft review written by me. Help me paraphrase it to improve grammar and clarity. You must preserve all technical arguments, \\ criticisms and core meaning without alteration.  \\
  Draft review: \\
  \{HUMAN\_REVIEW\} \\
  Start your response directly with the review text. Do not include any introductory phrases or disclaimers (e.g., 'Sure, here is the review').
\end{quote}

\noindent Prompt Variant 3:
\begin{quote}
\small\ttfamily
I am a reviewer for \{CONFERENCE\}. The following is a draft review written by me. Help me write a polished review with improved grammar and clarity. You must preserve all technical arguments, criticisms and core meaning without alteration.  \\
  Draft review: \\
  \{HUMAN\_REVIEW\} \\
  Start your response directly with the review text. Do not include any introductory phrases or disclaimers (e.g., 'Sure, here is the review').
\end{quote}

\end{document}